\documentclass[letterpaper]{article} 
\usepackage{aaai25}  
\usepackage{times}  
\usepackage{helvet}  
\usepackage{courier}  
\usepackage[hyphens]{url}  
\usepackage{graphicx} 
\usepackage{natbib}  
\usepackage{caption} 
\usepackage{subcaption}
\usepackage{algorithm}
\usepackage{algorithmic}
\usepackage{booktabs}
\usepackage{multirow}
\usepackage{amsmath}
\usepackage{rotating}

\usepackage{newfloat}
\usepackage{listings}
\DeclareCaptionStyle{ruled}{labelfont=normalfont,labelsep=colon,strut=off} 
\floatstyle{ruled}
\newfloat{listing}{tb}{lst}{}
\floatname{listing}{Listing}
\title{ABQ-LLM: Arbitrary-Bit Quantized Inference Acceleration for Large Language Models }

\author{
    Chao Zeng \equalcontrib, 
    Songwei Liu\equalcontrib \thanks{Project Leader.},
    Yusheng Xie\equalcontrib,
    Hong Liu,
    Xiaojian Wang, \\
    Miao Wei,
    Shu Yang,
    Fangmin Chen,
    Xing Mei \thanks{Corresponding author.}
}
\affiliations{
    \textsuperscript{\rm } ByteDance Inc.\\
    \{zengchaocs, cfangmin\}@gmail.com, \\
\{liusongwei.zju, xieyusheng.12, liuhong.10, wangxiaojian.wxj, weimiao.brandon, yang.1109, xing.mei\}@bytedance.com

}

\usepackage{bibentry}

\usepackage{tikz}
\newcommand*\circled[1]{\tikz[baseline=(char.base)]{
    \node[shape=circle,draw,inner sep=1pt,fill=black,text=white] (char) {#1};}}
    
\begin{document}

\maketitle

\begin{abstract}
Large Language Models (LLMs) have revolutionized natural language processing tasks. However, their practical application is constrained by substantial memory and computational demands. Post-training quantization (PTQ) is considered an effective method to accelerate LLM inference. Despite its growing popularity in LLM model compression, PTQ deployment faces two major challenges. First, low-bit quantization leads to performance degradation. Second, restricted by the limited integer computing unit type on GPUs, quantized matrix operations with different precisions cannot be effectively accelerated. To address these issues, we introduce a novel arbitrary-bit quantization algorithm and inference framework, ABQ-LLM. It achieves superior performance across various quantization settings and enables efficient arbitrary-precision quantized inference on the GPU. ABQ-LLM introduces several key innovations: (1) a distribution correction method for transformer blocks to mitigate distribution differences caused by full quantization of weights and activations, improving performance at low bit-widths. (2) the bit balance strategy to counteract performance degradation from asymmetric distribution issues at very low bit-widths (e.g., 2-bit). (3) an innovative quantization acceleration framework that reconstructs the quantization matrix multiplication of arbitrary precision combinations based on BTC (Binary TensorCore) equivalents, gets rid of the limitations of INT4/INT8 computing units. ABQ-LLM can convert each component bit width gain into actual acceleration gain, maximizing performance under mixed precision(e.g., W6A6, W2A8). Based on W2*A8 quantization configuration on LLaMA-7B model, it achieved a WikiText2 perplexity of 7.59 (2.17$\downarrow $ vs 9.76 in AffineQuant). Compared to SmoothQuant, we realized 1.6$\times$ acceleration improvement and 2.7$\times$ memory compression gain. Code will available at: \url{https://github.com/bytedance/ABQ-LLM}.
\end{abstract}

%

\section{Introduction}\label{sec:intro}

Recent advancements in large language models (LLMs) \cite{bubeck2023sparks, touvron2023llama, touvron2023llama2} have demonstrated impressive capabilities across various natural language benchmark, including reasoning \cite{clark2019boolq, clark2018think}, cognitive processing \cite{xu2023lvlm, hardy2023large}, and dialogue generation\cite{hu2023unlocking}. However, these models are characterized by a substantial number of parameters, posing significant challenges in terms of memory consumption and bandwidth \cite{zheng2024judging, kim2023squeezellm}.
Post-training quantization (PTQ) effectively reduces both computational and storage requirements. This technique significantly accelerates model inference by converting the weights and activation values of large language models (LLMs) from high-precision floating-point numbers to low-precision integer values for storage, and using efficient integer matrix multiplication operators to handle the bulk of matrix multiplication computations during inference. Currently, 80\% of the computation and parameter access in LLMs is concentrated on general matrix multiplication (GEMM) and vector multiplication (GEMV) operations, especially during autoregressive decoding, where all GEMM operations degrade into GEMV operations due to single-token generation. Consequently, the efficiency of GEMV computation and memory access directly determines the efficiency and power consumption of LLM inference.

To improve GEMM/GEMV memory access efficiency, LLM inference typically employs a quantized inference strategy. The current mainstream approach is weight-only quantization, where the kernel performs actual computation based on dequantized FP16 values. However, this approach offers limited performance improvement in highly parallel scenarios. To further enhance quantized inference performance, the industry is pursuing full quantization of both weights and activation values to reduce activation memory access and leverage higher computational power using quantized kernels, such as those from NVIDIA. However, current industry practices in weight and activation full quantization (WA full quantization) face several limitations. NVIDIA provides only a limited set of hardware-accelerated instructions\cite{lin2024awq, ashkboos2024quarot, zhao2024atom}, which constrains the design space for quantization algorithms. Other quantization combinations (e.g., W4A8 or W2A4) require type conversion to W8A8 or W4A4 during computation, leading to inefficiency\cite{lin2024qserve}. Furthermore, due to GEMV, additional padding calculations are required in scenarios with a batch size less than 8, resulting in inefficient matrix multiplication for W4A4 and W8A8. Finally, WA fully quantized models encounter significant challenges in low-bit quantization (e.g., W2A8, W2A6).

\begin{figure}[t]
\centering
\includegraphics[width=0.9\linewidth]{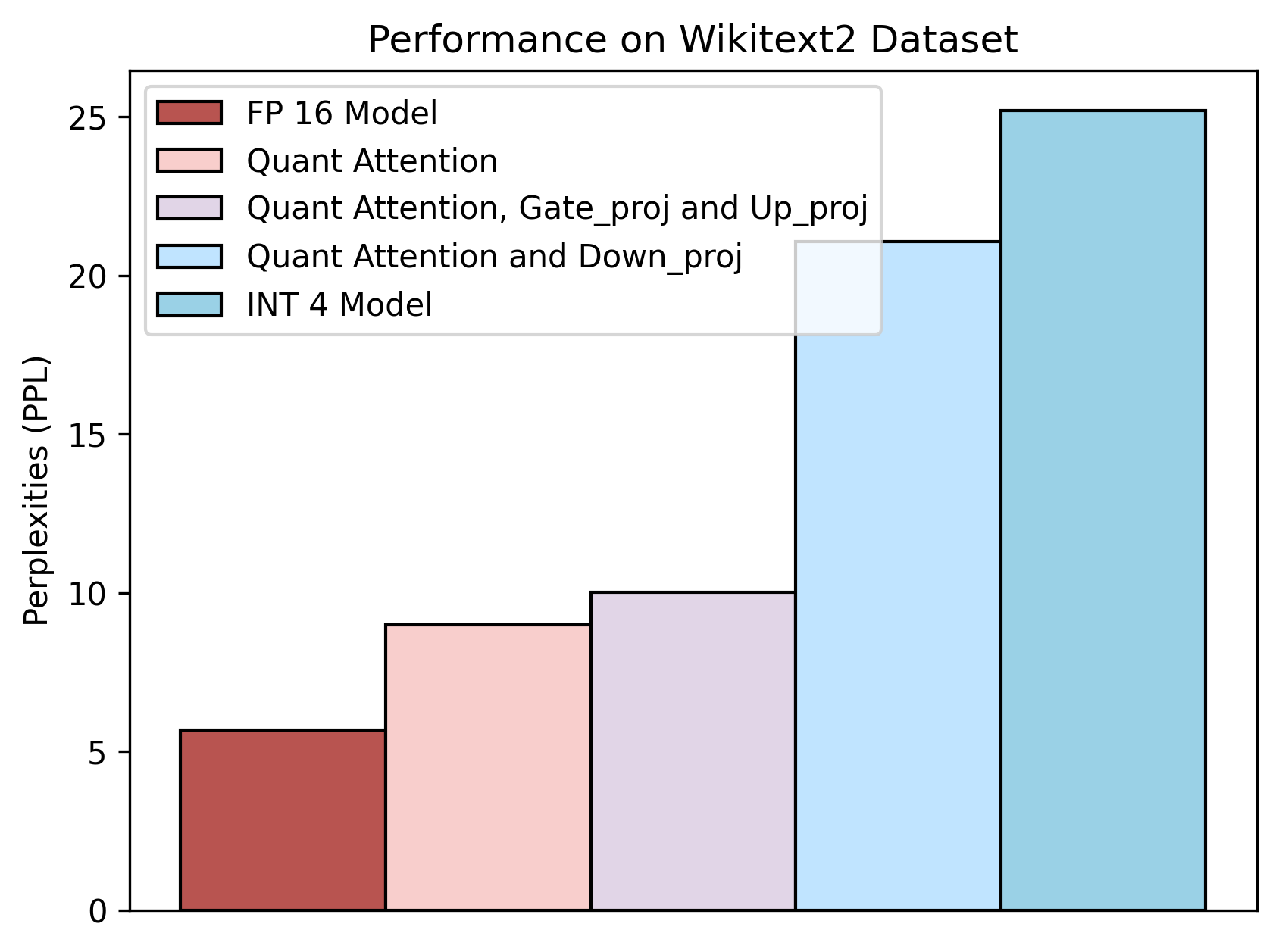}
\caption{Perplexities analysis (lower is better) on the Wikitext2 dataset of LLaMA-7B with different quantized modules.}
\label{fig:wikitext_ppl}
\end{figure}

\begin{figure}[t]
\centering
\includegraphics[width=\linewidth]{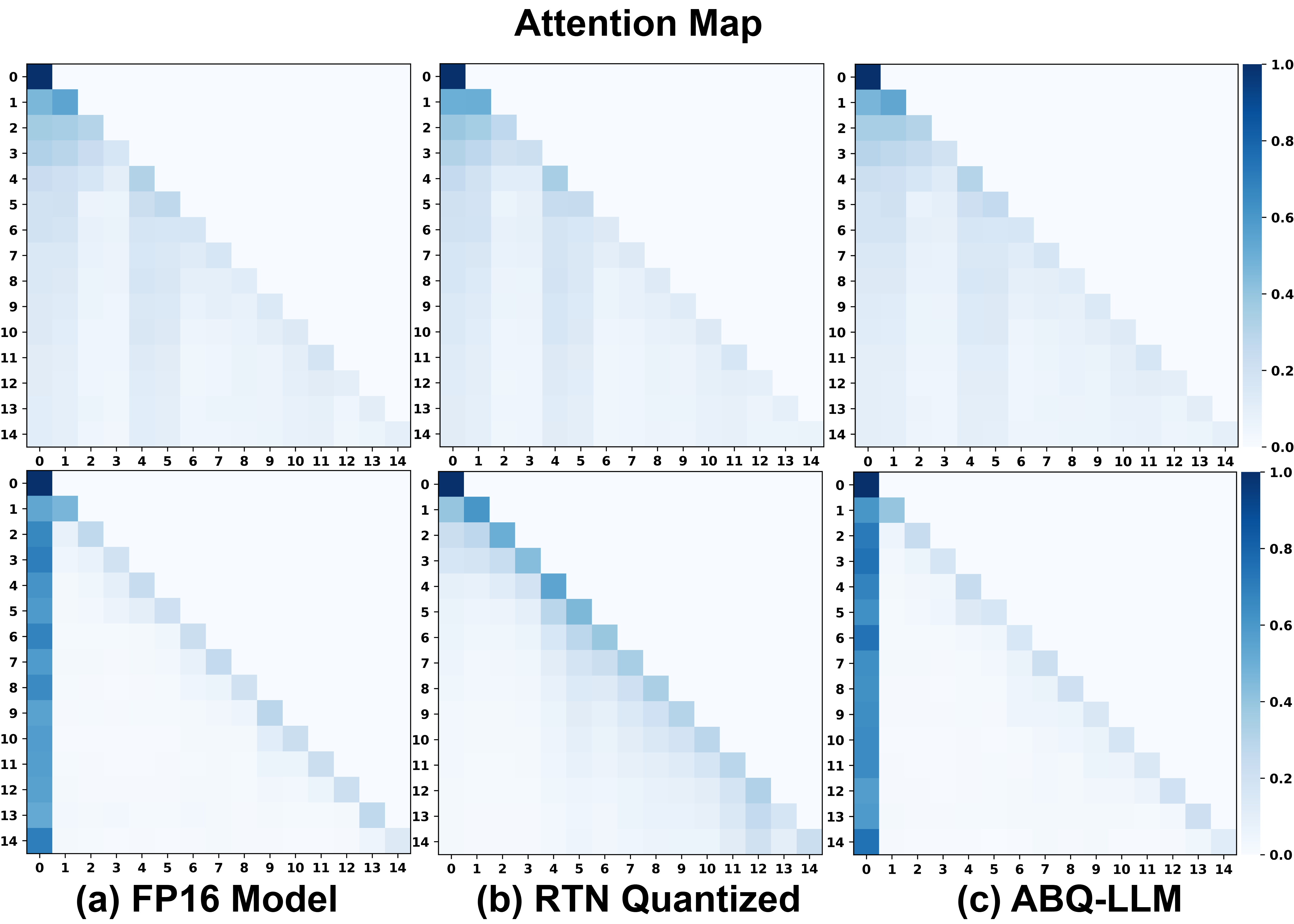}
\caption{Attention maps for the first block (above) and the final block (below) are shown.}
\label{fig:attention_distribution}
\end{figure}

In this paper, we introduce a novel quantization framework for PTQ, called ABQ-LLM. By examining the quantization sensitivity of components within the transformer block (Figure~\ref{fig:wikitext_ppl}) and the attention map before and after quantization (Figure~\ref{fig:attention_distribution}), we find the down\_proj linear layer and the attention map particularly sensitive to quantization. To address this, we propose a double cosine similarity distribution correction and an attention map distribution bootstrap for the output of down\_proj. This method calibrates the quantization constants and restores the model's performance at low bit-widths such as W6A6, W4A4 and W2A8. Additionally, we analyze performance degradation in low-bit quantization and address the asymmetric loss issue in low-bit representations like INT2 using the bit balance strategy. Finally, we implement customized software engine to support fully quantized inference of various precision combinations based on BTC equivalents, fully exploiting the advantages
of quantized models under mixed precision. Our contributions are summarized as follows:
\begin{itemize}
\item We propose a novel block-wise distribution correction and compensation scheme in the PTQ domain to mitigate the distribution discrepancy caused by full quantization of weights and activations, thereby improving model performance at low bit-widths.
\item We address the problem of asymmetric loss at low bit-widths, such as INT2, and significantly improve INT2 quantization performance using the bit balance strategy, enhancing model performance under the INT2 quantization configuration.
\item We propose a software engine which achieves quantization freedom for the first time in the LLM field. It eliminates the limitations of INT4/INT8 computational units, and effectively avoids the GEMV problem. Under the LLaMA-7B W2A8 configuration, it has 1.6$\times$ ultimate acceleration compared to SmoothQuant, achieving SOTA performance.
 \end{itemize}

\section{Related Work}
LLM quantization can be broadly divided into weight-only quantization and weight-activation quantization. 

\noindent \textbf{Weight-only quantization.} To alleviate computational burdens, some studies focus on weight-only quantization. LLM.int8() \cite{dettmers2022gpt3} achieves accurate INT8 quantization by retaining significant channels. GPTQ \cite{frantar2022gptq} uses Hessian-based error compensation to reduce quantization errors in LLMs, enabling 3-bit quantization. AWQ \cite{lin2024awq} and OWQ \cite{lee2024owq} significantly enhance quantized model performance by considering the impact of activation outliers on weight quantization. Methods like QuIP\cite{chee2024quip}, QuIP\# \cite{tseng2024quip}, and AQLM\cite{egiazarian2024extreme} facilitate 2-bit quantization through learnable codebooks or additional fine-tuning. Approaches such as \cite{dettmers2023spqr,shang2023pb,huang2024billm} improve PTQ performance through unstructured mixed-precision fine-grained weight grouping. Additionally, research such as \cite{dettmers2024qlora,xu2023qa,arshia2022peqa,bondarenko2024low} employs efficient parameter fine-tuning (PEFT) techniques to compress weights through fine-tuning.

\noindent \textbf{Weight-activation quantization.} Weight-activation quantization differs from weight-only quantization by quantizing both weights and activation (including KV caches) to accelerate LLM inference. The main challenge in quantizing activation is handling outliers, which can cause significant quantization errors. To address this issue, ZeroQuant\cite{yao2022zeroquant} proposes a fine-grained, hardware-friendly quantization scheme for weights and activation. SmoothQuant\cite{xiao2023smoothquant} shifts the quantization difficulty from activation to weights through mathematically equivalent transformations, achieving W8A8 quantization. \cite{shao2023omniquant,ma2024affinequant,hu2024llm} enhances performance by training quantization parameters. Limited by GPU platform instruction limitations, these jobs can only use W8A8 to perform actual inference, even if they achieve lower quantization bit-widths (e.g., W6A6).
\section{Method}
In this section, we provide a detailed introduction to our ABQ-LLM. We first describe the distribution correction and bit balance strategy and then introduce our arbitrary-bit inference framework.

\begin{figure}[t]
\centering
\includegraphics[width=\linewidth]{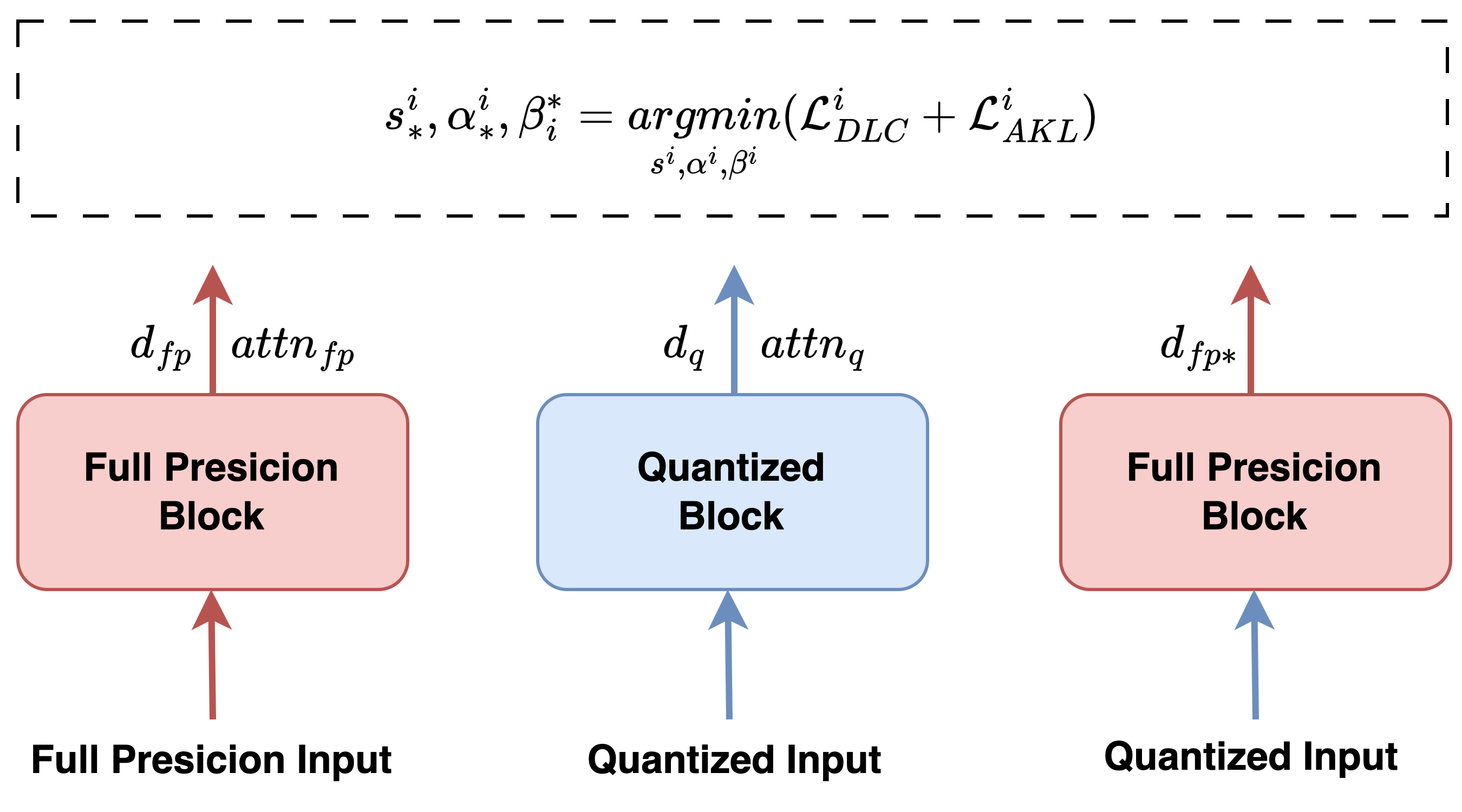}
\caption{An overview of our ABQ-LLM. ABQ-LLM use our DLC loss and AKL loss to update learnable parameters.}
\label{fig:abq-llm}
\end{figure}
\subsection{Preliminary}

\cite{xiao2023smoothquant} achieves WA full quantization by scaling activation outliers, but this increases the range variability of weights, making weight quantization more sensitive. Conversely, \cite{lin2024awq} optimizes weight quantization by scaling weights, which significantly increases the diversity of activation, complicating activation quantization. These approaches highlight the drawbacks of manually setting scaling balance factors between activation and weight, making it challenging to achieve a perfect balance. To address this issue, we introduce a distribution correction-guided scaling method. Following \cite{shao2023omniquant} approach, we set the balance vectors between weights and activation as learnable parameters and add a learnable clipping parameter for weight. By employing our distribution correction and bit balance strategy to optimize model performance, our objectives are as follows:

\begin{equation}\label{eq:balance} 
\begin{aligned}
\scalebox{0.9}{$
\underset{s, \alpha ,\beta }{\text{argmin}}\left \| WX-Q(\text{clip}(W)\cdot \text{diag}(s))Q(\text{diag}(s)^{-1} \cdot X) \right \|,
$}
\end{aligned}
\end{equation} 

\noindent where $W$ and $X$ are full-precision weight and activation, $Q(\cdot)$ denotes the quantizer of weight and activation, $\text{clip}(\cdot)$ denotes the clipping operation, $s$ is the scale factor, and letting $W_{max} = \alpha \text{max}(W)$ and $W_{min} = \beta \text{min}(W)$ to control the clipping range of the weight.

\subsection{Improving Quantization by Distribution Correction}
We observed significant variations in sensitivity across different layers of LLM models during quantization, with some layers having a critical impact on quantization performance. To validate this, as shown in Figure~\ref{fig:wikitext_ppl}, we quantified various components of the LLaMA-7B model under weight-activation full quantization. While quantizing the gate\_proj and up\_proj layers in mlp and attention resulted in only minor performance degradation, quantizing the down\_proj linear layer caused a substantial performance drop. This indicates that addressing down\_proj quantization is crucial for performance recovery. Further analysis revealed that the primary cause of performance degradation due to down\_proj quantization is the quantization of down\_proj activation. At low bit-widths such as INT4, INT3, and INT2, the limited representation range causes a significant shift in the model distribution compared to full precision. As illustrated in Figure~\ref{fig:abq-llm}, during the block-wise quantization calibration process, we apply a \underline{d}ouble \underline{l}ogarithm of \underline{c}osine similarity loss on the output of down\_proj to correct the distribution of the quantized model. The loss function called DLC loss $\mathcal L^{i}_{DLC}$:
\begin{equation}\label{eq:down_proj_correct}
\begin{aligned}
\mathcal L^{i}_{DLC} = -\text{log}(\frac{d^{i}_{q}\cdot d^{i}_{fp}}{\left \| d^{i}_{q} \right \| \left \| d^{i}_{fp} \right \| } )-\text{log}(\frac{d^{i}_{q}\cdot d^{i}_{fp^{*}}}{\left \| d^{i}_{q} \right \| \left \| d^{i}_{fp^{*}} \right \| } ) ,
\end{aligned}
\end{equation}
\noindent where $d^{i}_{q}$ represent the quantized output of the $i$-th transformer block, $d^{i}_{fp}$ represent the full-precision output of the $i$-th transformer block, and $d^{i}_{fp^{*}}$ represent the full-precision output of the $i$-th transformer block, with its input originating from the quantized output of the $(i-1)$-th transformer block.

Additionally, we conducted an analysis of the cosine similarity between activation at the input and output of decoder blocks in the LLaMA-7B model. The results revealed significant differences in similarity for the initial and final blocks, indicating their considerable impact on model inference performance. In response, we applied distribution compensation vector to the down\_proj layers of these blocks to address and correct the distribution discrepancies using Eq.~\eqref{eq:down_proj_com}.
\begin{equation}\label{eq:down_proj_com}
\begin{aligned}
W_{q} = \text{clamp}(\lceil \frac{W+\gamma ab^{\top} }{\bigtriangleup }  \rfloor + z, 0, 2^{n}-1 ),
\end{aligned}
\end{equation}

\noindent where $\lceil \cdot \rfloor$ denotes round operation, $n$ represents the target bit-width, $\bigtriangleup$ denotes the step-size, and $z$ is the zero-point. $W_{q}$ and $W$ denote the quantized and full-presicion weight, respectively. The vectors $a$ and $b$ are distribution compensation vectors, where $\gamma = 1$ indicates compensation is performed, and $\gamma=0$ indicates no compensation.


\begin{figure*}[!htb]
\centering
\begin{subfigure}[t]{0.60\linewidth}
\centering
\includegraphics[width=\linewidth]{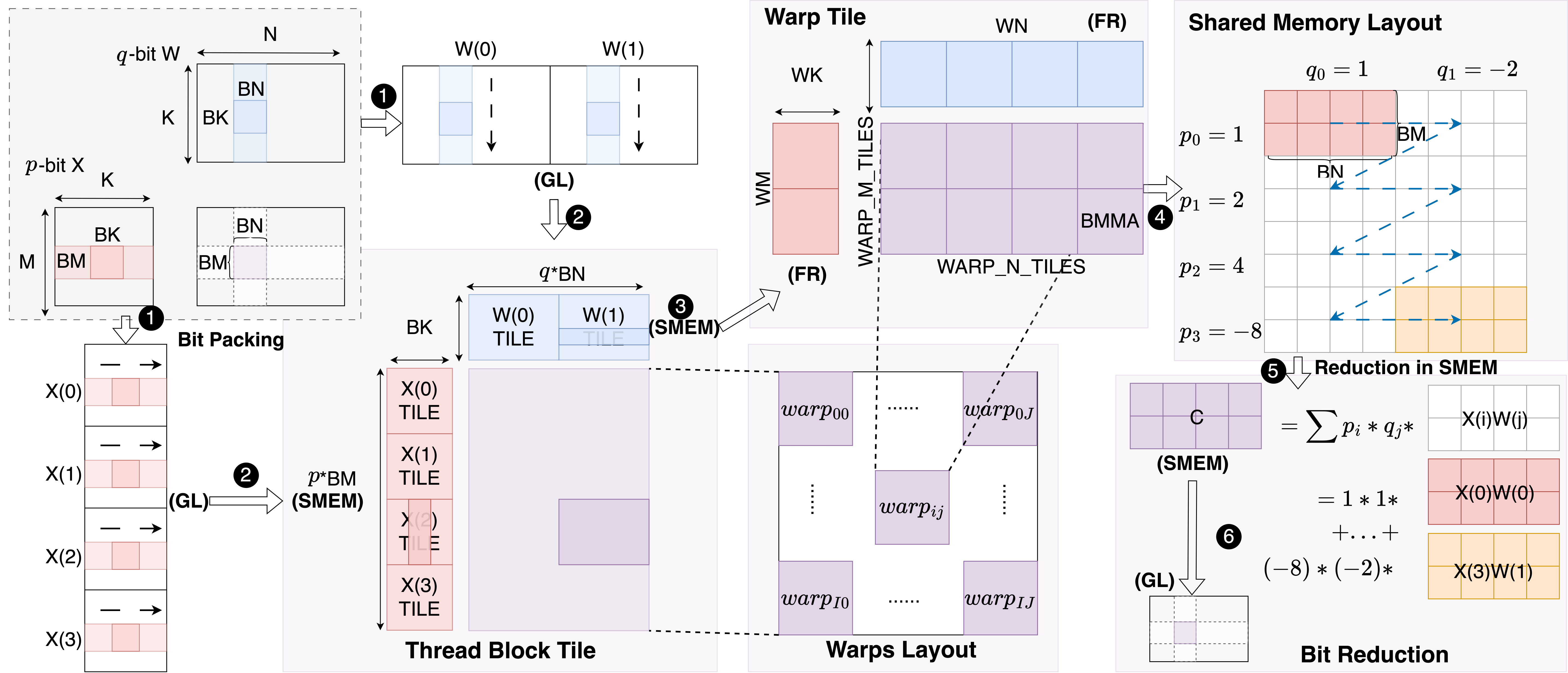}
\caption{}
\end{subfigure}%
\hfill
\begin{subfigure}[t]{0.36\linewidth}
\centering
\includegraphics[width=\linewidth]{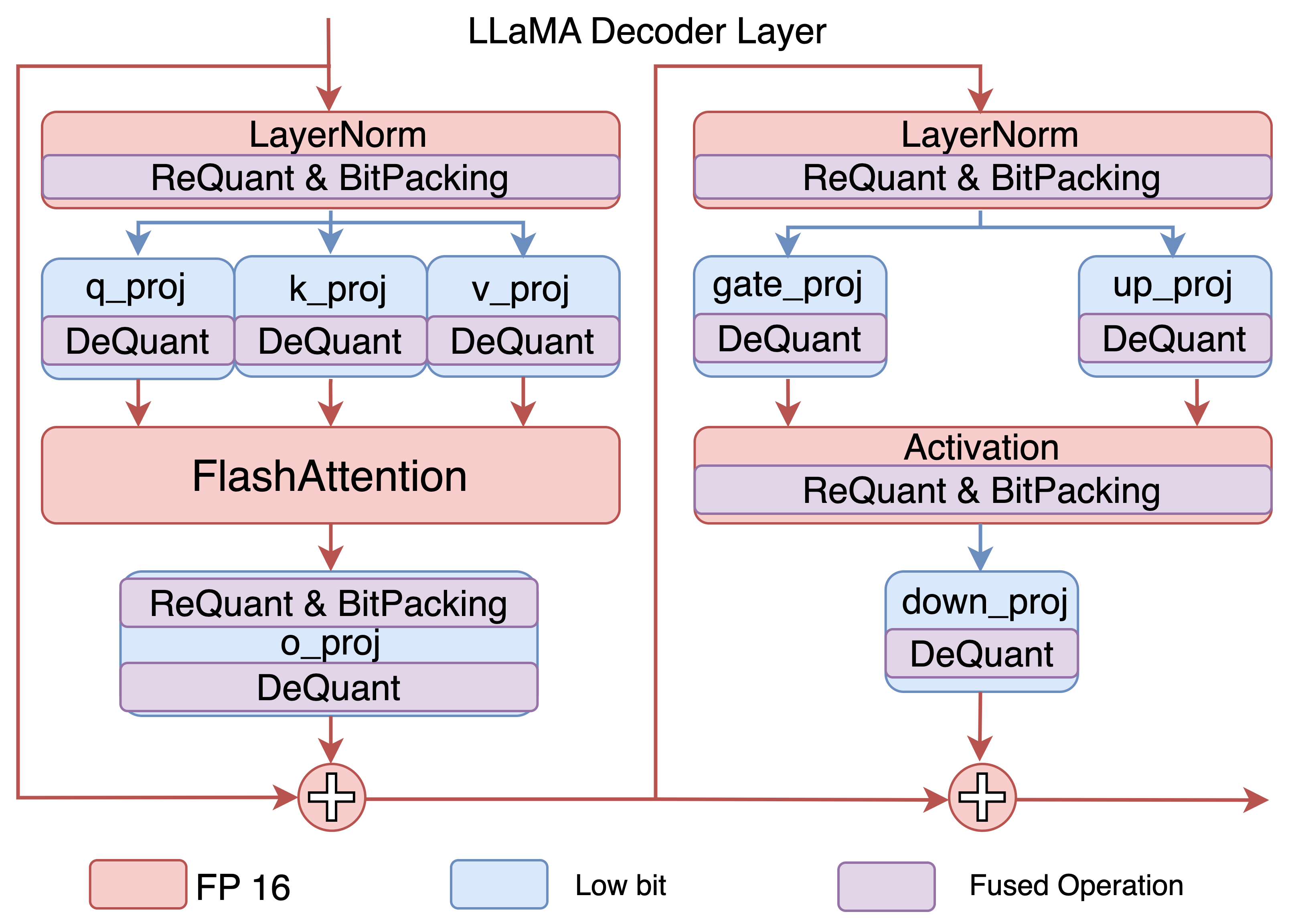}
\caption{}
\end{subfigure}
\caption{(a) System Overview of Custom  Software Engine. 
\textbf{p} and \textbf{q} represent the quantization bit width of input X and weight W.  Data flows are carefully designed to support efficient computation: global memory(\textbf{GL}) $\rightarrow$ shared memory(\textbf{SMEM})  $\rightarrow$ fragment(\textbf{FR})  $\rightarrow$ shared memory(\textbf{SMEM})  $\rightarrow$ global memory(\textbf{GL}). (b) ABQ-LLM’s calculation diagram for a Transformer block. \textbf{ReQuant} and \textbf{DeQuant} represent online quantization and dequantization operations respectively, and \textbf{BitPacking} represents the online layout transformation for activation.}
\label{fig:mmba}
\end{figure*}

To enhance the performance of the quantized model, we analyzed the changes in Attention Map distribution before and after quantization, as shown in Figure~\ref{fig:attention_distribution}. In the full-precision model, attention is heavily focused on the first token, highlighting its key role in guiding text generation, consistent with LLM-QAT\cite{liu2023llm} findings. However, quantization disrupts this attention pattern, diminishing focus on the first token. To address this and restore the model's attention during quantization, we introduced \underline{a}ttention-aware \underline{KL} divergence to reconstruct the attention map.
\begin{equation}\label{eq:attention_aware_correct} 
\begin{aligned}
\scalebox{0.9}{$
\mathcal L^{i}_{AKL} = D_{KL}(attn^{i}_{q}\parallel attn^{i}_{fp}) + D_{KL}(attn^{i}_{fp}\parallel attn^{i}_{q}),
$}
\end{aligned}
\end{equation}
\noindent where $attn^{i}_{q}$ denotes the quantized attention map output of the $i$-th transformer block, while $attn^{i}_{fp}$ refers to the full-precision attention map output of the same block.

In the end, we combined DLC loss and AKL loss, and our final optimization goal is:
\begin{equation}\label{eq:loss}
\begin{aligned}
s^{i}_{*}, \alpha^{i}_{*}, \beta^{*}_{i} = \underset{s^{i}, \alpha^{i} ,\beta^{i} }{\text{argmin}} (\mathcal L^{i}_{DLC} + \mathcal L^{i}_{AKL} ),
\end{aligned}
\end{equation}
\noindent where $s^{i}_{*}$, $\alpha^{i}_{*}$ and $\beta^{*}_{i}$ are the parameters of the $i$-th transformer block after calibration. When the distributions of the quantized output and the full-precision output match, we have a loss close to 0, which effectively guides the quantization process.

\subsection{Bit Balance Strategy}
Typically, pre-trained LLM model weight exhibit near-normal distribution, characterized by symmetry. Using Q-Q plots (Quantile-Quantile Plots), we confirmed the strong symmetry in the weight distribution of pre-trained models (see Appendix \ref{a:bit_analyse}). However, in standard INT2 quantization, the numerical representation is limited to four values, with symmetric quantization ranges of \{-2, -1, 0, 1\} or \{-1, 0, 1, 2\}, disrupting the original symmetric weight distribution (see Appendix \ref{a:bit_analyse}). This asymmetry leads to significant performance degradation, as shown in Table~\ref{tab:bit-balance-aba}, where performance drops by 0.46 from W4A16 to W3A16 and by 5.19 from W3A16 to W2A16, indicating a sharp decline. To address this asymmetry impact on LLM quantization, we adopted the bit balance strategy like \cite{li2016ternary,ma2024era}, extending the INT2 symmetric quantization space to \{-2, -1, 0, 1, 2\}. This modification restored model performance to 7.50, which is within a reasonable range compared to W3A16.

\subsection{Custom Software Engine}
\noindent \textbf{Reconstructing Arbitrary Bit Computation.} To support W1A1 quantization, NVIDIA introduced INT1 TensorCore in Turing and later architectures to provide hardware support. However, W1A1 quantization has not been widely applied due to significant performance degradation. Through mathematical analysis of quantized matrix multiplication, we find that any combination of quantization can be decomposed into a superposition of 1-bit matrix multiplications. Assuming the weight $W$ of a particular neural network layer are quantized to \textit{q} bits and the input activation value $X$ is quantized to \textit{p} bits, the matrix multiplication of $W$ and $X$ results in a 32-bit output $Y=WX$. The key is to observe that the scalar values at any position in W and X can be decomposed into a series of 1-bit scalar numbers. Scalar operations with any combination of precision can be decomposed into 1-bit operations and shift operations. For example, a 2-bit x can be expressed as:
\begin{equation}
\begin{aligned} 
x = x^{1}x^{0} , \text{where} \quad x^{i} \in \text{INT1},
\end{aligned}
\end{equation}
where $x^{1} = (x \gg 1)\&1$, $x^{0}=(x \gg 0)\&1$. We use $OP(a,b)$ to denote a computational operation where the input is 1-bit data and the output is 32-bit. Thus, the original scalar-level arbitrary precision computation $wx$ can be represented as:
\begin{equation}
\begin{aligned} 
wx=w*(x^{1}x^{0}) = OP(w,x^{1}) *2 + OP(w,x^{0}).
\end{aligned}
\end{equation}

The above procedure can be generalized to any combination of matrix multiplications with bit-widths \textit{p} and \textit{q}. The detailed formulas are shown in the Appendix~\ref{a:abc}. Using these transformations, we decompose the operation of arbitrary quantized combinations into a superposition of 1-bit matrix multiplications, enabling the underlying layer to invoke high-computing instruction implementations. 

\begin{table}[t]
\centering
\begin{tabular}{cccc}
\toprule
Model                     & Bits   & WikiText2 & C4    \\ \midrule
\multirow{4}{*}{LLaMA-7B} & W4A16  & 5.83      & 7.29  \\
                          & W3A16  & 6.29      & 8.01  \\
                          & W2A16  & 11.48     & 15.74 \\
                          & W2*A16 & 7.50      & 9.86  \\ \bottomrule
\end{tabular}
\caption{Performance comparison of LLaMA-7B under W4, W3, and W2 quantization configurations. * denotes the use of the bit balance strategy.}
\label{tab:bit-balance-aba}
\end{table}

\noindent \textbf{Engine Implementation.} NVIDIA GPU has many processing elements called Streaming Multiprocessors (SMs) and uses a large number of threads to perform computing tasks in parallel. Threads are structured into thread blocks, which become the smallest scheduling execution unit on SMs. Therefore, the computation target is decomposed and mapped to each thread block, called 
\textbf{Thread Block Tile}, to achieve parallel computing. As shown in Figure \ref{fig:mmba}(a), for a GEMM task of shape M×N×K, each thread block is responsible for computing a BM×BN output block, which is decomposed into $\frac{K}{BK}$ sub-GEMM tasks of shape BM×BN×BK. Our engine converts quantized matrix multiplications with bit widths configured as \{p, q\} into special accumulations of p*q binarized matrix multiplications, so the true calculation task of thread block tile is p*BM × q*BN. \circled{1} First, in order to improve memory access continuity, we propose \textbf{BitPacking} strategy to decompose the quantized tensor into $n$ binary matrices, where $n$ is the quantization bit width. Taking input X as an example, this means that its bit perspective layout changes from [M, K, p] to [p, M, K]. All threads within a thread block share the same shared memory space. Within each thread block, threads are further organized into a set of warps, with each warp consisting of 32 consecutive threads. \circled{2} Next, warps collaborates to load the A matrix (p*BM × BK) and B matrix (BK × q*BN) data required for thread block tile  calculation from GL and caches them in SMEM. Thanks to BitPacking, the process of reading p BM*BK single-bit row-major tiles and writing p*BM*BK bits SMEM is efficient and continuous.  \circled{3} Subsequently, thread block contains multiple warps, so thread block tile can be further decomposed into \textbf{Warp Tile} to achieve warp-level parallelism, and the computing tasks of each warp are WM× WN. In the calculation preparation stage, the A matrix (WM×WK, row-major) and B matrix (WK×WN, col-major) are independently loaded from SMEM to FR.  Then, the calculation is decomposed into WM\_TILES*WARP\_N\_TILES TensorCore MMA(matrix-multiply-accumulate). Since A and B are binarized matrices, we actually use Binary TensorCore MMA (BMMA), which has a computing power 8 times and 4 times higher than INT8 and INT4 TensorCore respectively. \circled{4} All warps collaboratively complete the Thread Block Tile calculation, and the results are stored in the c fragment of each warp. Therefore, each warp needs to independently write the calculation results back to SMEM. \circled{5} As shown in Eq.~\eqref{formula:matrix}, output tile(p*BM × q*BN) is globally reduced to obtain a final result(BM × BN), where each BM × BN sub-Tile needs to be multiplied by a certain scaling factor. We call this process \textbf{Bit Reduction}.
\circled{6} As the final step, warps collaboratively load the final result from SMEM and write back to the target location in GL. 

We implement the above calculation process as a GPU Kernel, called ABQKernel. As shown in Figure~\ref{fig:mmba}(b), ABQKernel is used to replace all gemm operations in the decoder layer, and assists with necessary BitPacking, quantization, and dequantization operations to achieve arbitrary quantization inference of the LLaMA model. We carefully manages the overhead of quantization operators by fusing them into existing operators and weight BitPacking is implemented offline for increased efficiency.

\noindent \textbf{GPU Kernel Optimization.}  When M=1, the GEMM problem of shape MxNxK transforms into a GEMV problem, shifting from computation-intensive to memory-intensive, which becomes a performance bottleneck for model inference. When using ordinary TensorCore for accelerated computation, the dimensions of M are usually chunked in groups of 8, requiring padding if M$<$8, leading to 87.5\% redundant computation. Thanks to the revolutionary reconstruction of computing and BitPacking strategy, for the $W_{q}A_{p}$ configuration, the actual computing task undertaken by ABQKernel is p*M × q*N × K. The expansion of the M dimension can effectively reduce the redundant calculations when calling TensorCore, and even when p*M $>=$ 8 and p*M $\%$ 8 = 0, padding can be completely avoided. We call the above optimization strategy \textit{GEMV Elimination}. In addition, \textit{Computational and Pipeline Optimization}, \textit{ Auto Kernel Search}, and \textit{Bank Conflicts Elimination} are also applied. For details, see Appendix \ref{gpu_opt}.

\section{Experiments}

\begin{table*}[!ht]
\centering
\begin{tabular}{cccccccccc}
\toprule
\multirow{2}{*}{Bits} & \multirow{2}{*}{Method} & \multicolumn{2}{c}{LLaMA-7B}   & \multicolumn{2}{c}{LLaMA-13B}  & \multicolumn{2}{c}{LLaMA-2-7B} & \multicolumn{2}{c}{LLaMA-2-13B} \\ \cmidrule(l){3-10} 
                      &                         & WikiText2     & C4             & WikiText2     & C4             & WikiText2     & C4             & WikiText2      & C4             \\ \midrule
\multirow{4}{*}{W6A6} & SmoothQuant             & 6.03          & 7.47           & 5.42          & 6.97           & 6.20          & 7.76           & 5.18           & 7.67           \\
                      & OmniQuant               & 5.96          & 7.43           & 5.28          & 6.84           & 5.87          & 7.48           & 5.14           & 6.74           \\
                      & I-LLM                   & 5.84          & 7.32           & 5.23          & 6.79           & 5.68          & 7.27           & 5.10           & 6.74           \\ 
                 & \textbf{ABQ-LLM}        & \textbf{5.81} & \textbf{7.27}  & \textbf{5.21} & \textbf{6.77}  & \textbf{5.63} & \textbf{7.21}  & \textbf{5.00}  & \textbf{6.64}  \\ \midrule
\multirow{5}{*}{W4A4} & SmoothQuant             & 22.25         & 32.22          & 40.05         & 47.18          & 83.12         & 77.27          & 35.88          & 43.19          \\
                      & OmniQuant               & 11.26         & 14.51          & 10.87         & 13.78          & 14.26         & 18.02          & 12.30          & 14.55          \\
                      & AffineQuant             & 10.28         & 13.64          & 10.32         & 13.44          & 12.69         & 15.76          & 11.45          & 13.97          \\
                      & I-LLM                   & 9.10          & 12.33          & 7.99          & 10.96          & 10.55         & 12.92          & 9.76           & 12.57          \\ 
                & \textbf{ABQ-LLM}        & \textbf{8.63} & \textbf{12.10} & \textbf{7.69} & \textbf{10.90} & \textbf{9.31} & \textbf{12.85} & \textbf{8.62}  & \textbf{11.47} \\ \midrule
\multirow{4}{*}{W2A8}  & OmniQuant   &    15.70     &      26.44   &     13.50    &    19.01  &     37.95   &   103.39   &    21.74            &     31.72    \\
                      & AffineQuant    &    \textbf{9.76}   & 15.52   &  9.21  &  12.55 &  1483       &   4688    &   \textbf{12.30}   &  29.32             \\
                      & I-LLM     &  14.08   &  18.89   &  11.80    &  16.19    & 123.93 &         200.54  &  25.74   & 40.59   \\
             & \textbf{ABQ-LLM}        & 11.35   & \textbf{15.41}      & \textbf{9.20}     & \textbf{12.48}      & \textbf{13.47}   & \textbf{17.82}      & 13.24      & \textbf{18.07}      \\ \midrule
            W2*A8  & \textbf{ABQ-LLM}        &   \textbf{7.59}      & \textbf{10.00}      & \textbf{6.49}     & \textbf{8.53}      & \textbf{7.85}     & \textbf{10.33}      & \textbf{6.65}      & \textbf{10.01} 
                      \\ \bottomrule
\end{tabular}
\caption{Weight-activation quantization perplexities (lower is better) comparison of quantized LLaMA and LLaMA-2 models. * denotes the use of the bit balance strategy. More results can be found in at Appendix~\ref{quantization_all_result} Tables~\ref{weight-only}, \ref{tab:other-setting}.}
\label{weight-activation}
\end{table*}

\subsection{Experimental Setup}
\noindent \textbf{Baseline.} For weight-only quantization, we compare our approach with GPTQ\cite{frantar2022gptq}, AWQ\cite{lin2024awq}, OmniQuant\cite{shao2023omniquant}, and AffineQuant\cite{ma2024affinequant}. For weight-activation quantization, we benchmark our method against SmoothQuant\cite{xiao2023smoothquant}, OmniQuant\cite{shao2023omniquant}, and I-LLM\cite{hu2024illm}.

\noindent \textbf{Models and Datasets.}  We primarily evaluate our method using LLaMA (7B-13B) \cite{touvron2023llama} and LLaMA-2 (7B-13B) \cite{touvron2023llama2}in this paper. Following previous work\cite{shao2023omniquant,ma2024affinequant}, we evaluate the quantized models by reporting the perplexity of language generation experiments on WikiText2\cite{merity2016pointer} and C4\cite{raffel2020exploring}. To assess performance on zero-shot tasks, we select several popular benchmarks including PIQA\cite{bisk2020piqa}, ARC\cite{clark2018think}, BoolQ\cite{clark2019boolq}, HellaSwag\cite{zellers2019hellaswag}, and Winogrande\cite{sakaguchi2021winogrande} using the lm-evaluation-harness\cite{gao10256836framework}.



\noindent \textbf{Calibration} We initialize the balance vectors for weights and activations following \cite{xiao2023smoothquant}, with the learnable clipping parameter for weights set to 1. For distribution correction compensation vectors, we set $a$ as an all-ones vector and $b$ as an all-zeros vector, ensuring $ab^{\top}$ starts at 0. Using the AdamW optimizer \cite{adamw} with no weight decay, we set learning rates of 5e-3 for balance vectors and 1e-2 for the clipping parameter and vector compensation vector. Calibration data includes 128 randomly selected 2048-token segments from WikiText2. The calibration process, conducted on an NVIDIA A800-40G GPU, utilized a batch size of 1 and spanned 20 epochs. For activation and KV Cache we perform per-token quantization, and for weight we perform per-channel quantization. By default, activation and KV cache use the same quantization bit.

\begin{figure}[t]
\centering
\includegraphics[width=0.95\columnwidth]{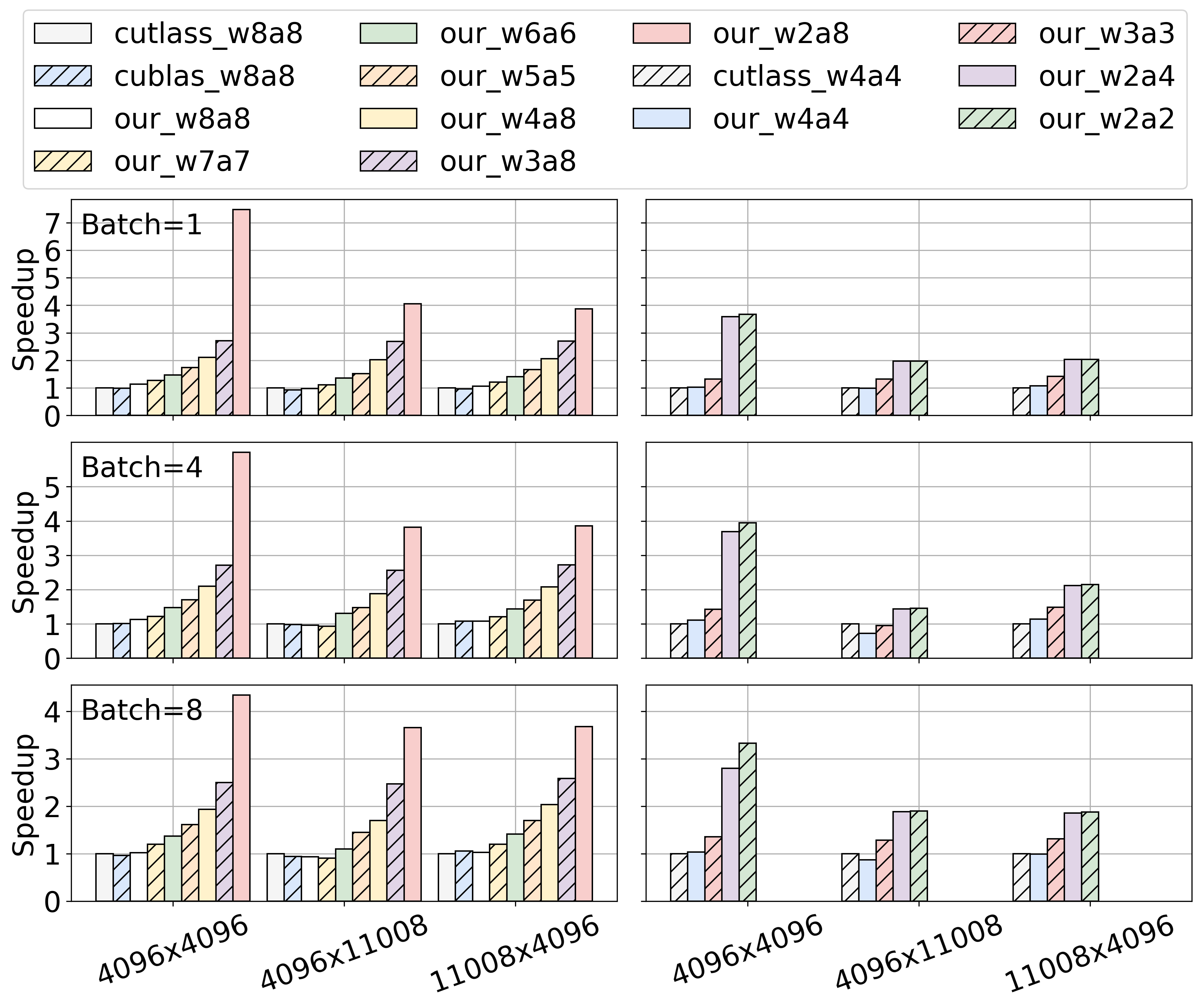}
\caption{The GEMV speedup comparison of our ABQKernel, CUTLASS (W4A4/W8A8), and cuBLAS (W8A8) in RTX 3070. The left side is compared against W8A8 and the right side against W4A4.
More result in RTX 4080 can be found in Appendix~\ref{gpu_opt} tables~\ref{tab:3070}, \ref{tab:4080}.}
\label{fig:kernel_speed}
\end{figure}

\subsection{Experiments on Language Generation Tasks}
Language generation capability is central to large language models (LLMs). To validate our extraordinary performance in the challenging quantization task, we first compare perplexity, a crucial metric for language generation, with the baseline. As shown in Table~\ref{weight-activation}, ABQ-LLM demonstrates outstanding performance across various quantization configurations, surpassing state-of-the-art methods such as AffineQuant and I-LLM. Notably, in the INT2 setting, the application of bit balance strategy yields significant improvements at minimal cost. The W2*A8 configurations substantially outperform the W2A8 configurations. Specifically, perplexity on WikiText2 and C4 datasets decreases by an average of 1.42 and 2.11 points, respectively, for W2*A8 compared to the W4A4. These findings validate the effectiveness of our distribution correction and bit balance strategy. The results of weight-only quantization are presented in Table~\ref{weight-only} at Appendix~\ref{quantization_all_result}. Additional results for full WA quantization are provided in Tables~\ref{tab:other-setting} at Appendix~\ref{quantization_all_result}.

\begin{table*}[t]
\centering
\begin{tabular}{cccccccccc}
\toprule
Model                        & Bits                  & Method      & PiQA           & ARC-e & ARC-c          & BoolQ & HellaSwag      & Winogrande & Avg.           \\ \midrule
\multirow{9}{*}{LLaMA-13B}  & \multirow{4}{*}{W4A4} & OmniQuant   & 69.69          & 47.30 & 33.10          & 62.84 & 58.96          & 55.80      & 54.37          \\
                             &                       & AffineQuant & 66.32          & 43.90 & 29.61 & \textbf{64.10} & 56.88          & 54.70      & 52.58          \\
                             & & I-LLM       & 67.95 & \textbf{48.15} & 34.47          & 62.29 & 63.13          & \textbf{59.98}      & 55.99   \\ 
                             &  & \textbf{ABQ-LLM}     & \textbf{71.82} & 47.60 & \textbf{35.67} & 63.52 & \textbf{64.31} & 57.54      & \textbf{56.74} \\ \cmidrule(l){2-10} 
                             & \multirow{4}{*}{W2A8} & OmniQuant   &  66.76  & 45.62  &  30.20     &    61.13   & 52.93   & 55.72   &  52.06   \\
                             & & AffineQuant &  71.00 &  46.70  &   \textbf{32.33}  & 62.23  &  58.62  & \textbf{63.53}  &    55.73   \\
                             &  & I-LLM  &  67.46  &   43.73   &   29.69 & 62.41 &    53.37  &  55.09   &  51.95   \\ 
                    &  & \textbf{ABQ-LLM} & \textbf{72.03} & \textbf{46.72}& 31.74 &  \textbf{65.17}  &  \textbf{58.71} & 62.50 & \textbf{56.15} \\ \cmidrule(l){2-10}
                     &  W2*A8  & \textbf{ABQ-LLM} & \textbf{74.91} & \textbf{54.92}  & \textbf{38.65}  & \textbf{68.53}  & \textbf{68.21} & \textbf{66.54} &      \textbf{61.96} \\ \midrule
\multirow{9}{*}{LLaMA-2-13B} & \multirow{4}{*}{W4A4} & OmniQuant   &   67.08     & 45.66 &  32.25     &    63.73   &    58.39      &   54.61   &  53.62  \\
                             &     & AffineQuant &    67.68     &  46.63   &  32.85   & \textbf{65.90}   &   60.62  &    54.14        &      54.63      \\
                             &    & I-LLM       &   68.00 &     45.74  &  30.97    & 64.55   &  60.62    &       54.22     &  54.01 \\ 
                             &  & \textbf{ABQ-LLM}  & \textbf{69.04}    & \textbf{ 47.01}   &  \textbf{33.53}   &  64.74  & \textbf{62.70} & \textbf{54.38} &  \textbf{55.23} \\ \cmidrule(l){2-10} 
                             & \multirow{4}{*}{W2A8} & OmniQuant   & 62.67 & 38.80  & 28.41        &   62.11  &  49.04  &  51.69    &  48.78     \\
                             &    & AffineQuant &    61.31  &  38.51 & 26.96   &   62.04  & 41.92 &    50.74        &   46.91  \\
                             &  & I-LLM       &  61.86  &  38.67 & 26.45   & 62.17  & 43.30    & 51.85  &  47.38 \\ 
                             & & \textbf{ABQ-LLM}  & \textbf{64.30} & \textbf{40.19} & \textbf{29.78}  &  \textbf{63.18} & \textbf{49.58}  &  \textbf{52.17}  &   \textbf{49.87}  \\ \cmidrule(l){2-10}
                             & W2*A8 & \textbf{ABQ-LLM}  &  \textbf{73.50} & \textbf{49.79}  &  \textbf{35.15} & \textbf{70.15} & \textbf{67.45}  &  \textbf{58.87} &  \textbf{59.15}  \\ \bottomrule
\end{tabular}
\caption{Zero-shot accuracies (higher is better) comparison of quantized LLaMA and LLaMA-2 models. * denotes the use of the bit balance strategy. More results can be found in at Appendix~\ref{quantization_all_result} Tables~\ref{tab:llama-7b -zero-shot}, \ref{tab:llama-13b -zero-shot}, \ref{tab:llama-2-7b -zero-shot}, \ref{tab:llama-2-13b -zero-shot}.}
\label{tab:weight-activation-zero-shot}
\end{table*}

\subsection{Experiments on Zero-Shot Tasks}
To further validate our model, we compare zero-shot accuracy with the baseline method, as shown in Table~\ref{tab:weight-activation-zero-shot}. Our ABQ-LLM method outperforms the previous method in most cases. Notably, after applying the bit balance strategy, the performance of W2*A8 improves significantly by 7.50\% on average. Combining the performance in both language generation and zero-shot tasks, we conclude that ABQ-LLM achieves state-of-the-art results in handling challenging quantization tasks. See the Appendix~\ref{quantization_all_result} for more results and analysis of quantized configurations.

\subsection{Inference Engine Evaluation}
\noindent \textbf{Kernel Benchmark.} We evaluated the GEMV speedup of our ABQKernel across three matrix dimensions in LLaMA-7B and compared it with the quantization kernel provided by cuBLAS and CUTLASS. It is important to note that CUTLASS only supports W4A4 and W8A8, while cuBLAS supports only W8A8 for quantization operations. Our experiments were conducted on two different GPUs: the RTX 4080 and the RTX 3070. Figure~\ref{fig:kernel_speed} presents the results, showing that our ABQKernel achieve superior acceleration across all matrix sizes. Specifically, for special bit combinations such as W2A8 and W2A4, our ABQKernel significantly outperforms the baseline approaches, as cuBLAS and CUTLASS require conversion to W8A8 and W4A4 for computation. In the W2A8 configuration, our throughput is improved by 7.47$\times$ compared to W8A8 with CUTLASS and cuBLAS on dimensions (1, 4096) $\times$ (4096, 4096).

\begin{figure}[htb]
\centering
\includegraphics[width=0.95\columnwidth]{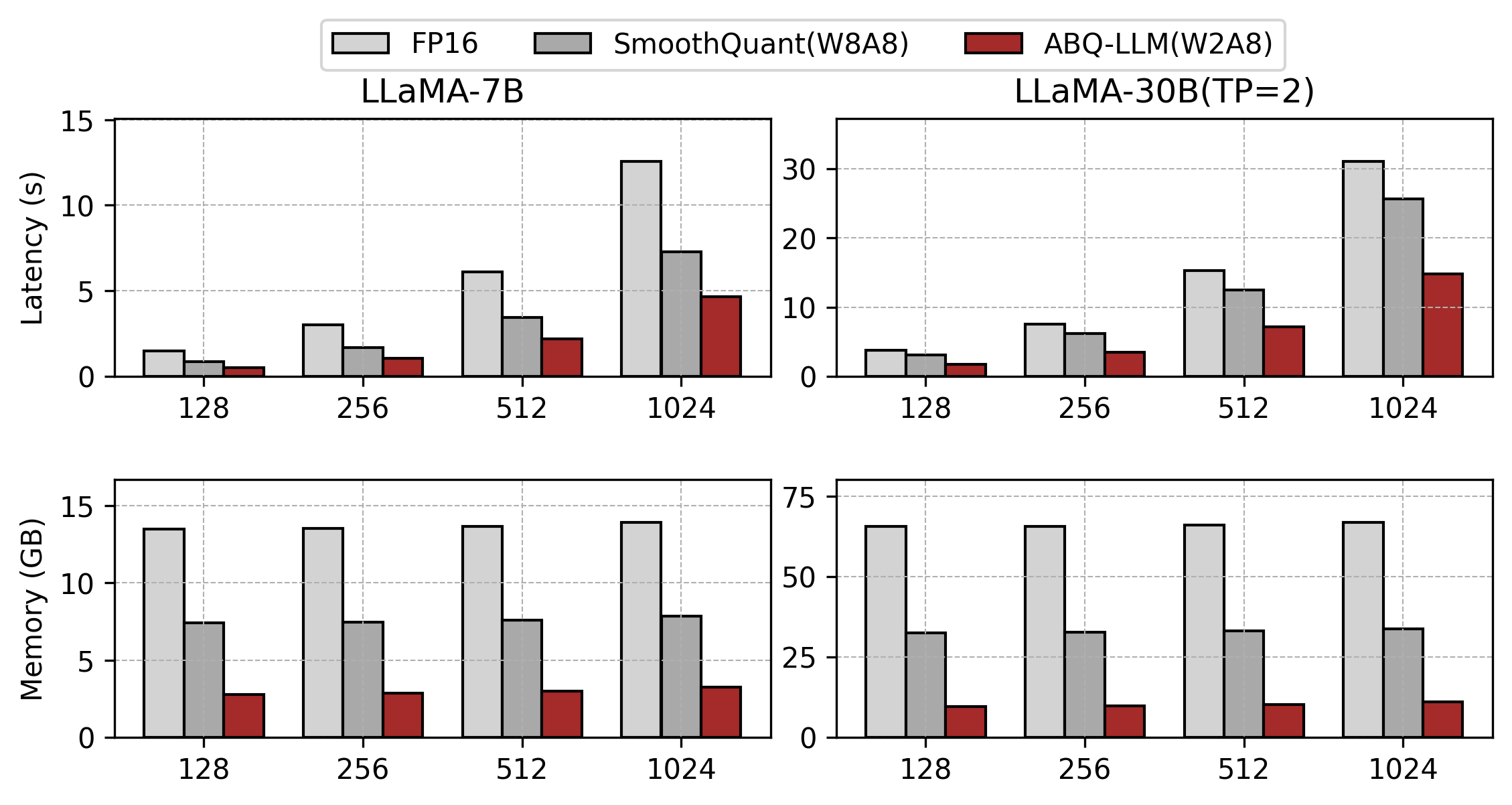}
\caption{Inference latency (top) and memory usage (bottom) of the FastTransformer implementation on NVIDIA A800-40GB GPU with a fixed input length of 15. More results can be found in at Appendix~\ref{gpu_opt} Table~\ref{tab:e2e}.}
\label{fig:e2e}
\end{figure}

\noindent \textbf{End-to-end throughput.}
As shown in Figure~\ref{fig:e2e}, we integrate our ABQKernel into FastTransformer and compare it with the FP16 implementation of FastTransformer and the INT8 implementation of SmoothQuant. Compared to FP16, our scheme achieves 2.95$\times$ inference acceleration and 4.8$\times$ memory compression gain, while requiring only 10GB of memory for inference on the LLaMA-30B model, which is less than the memory required for LLaMA-7B with FP16. Additionally, our scheme achieves 1.6$\times$ speedup and 2.7$\times$ memory compression gain over SmoothQuant, significantly outperforming current mainstream inference methods. This substantial improvement reduces the cost of LLM services and facilitates their practical deployment.


\noindent \textbf{Kernel Optimization Ablation.} Table 4 presents the impact of various optimization techniques on the inference latency of the kernel when performing the GEMV operation with dimensions (1, 4096) $\times$ (4096, 4096). 
Our ABQKernel significantly outperforms the CUTLASS W8A8 kernel when unoptimized. Additionally, by employing pipeline optimization, GEMV elimination and auto kernel search, we achieve a latency reduction by 7.47$\times$ and a corresponding increase in throughput by 7.47$\times$. These results significantly outperform CUTLASS.

\begin{table}[ht]
\centering
\begin{tabular}{ccc}
\toprule
Method           & Latency(us)$\downarrow $ & TOPS $\uparrow $ \\ \midrule
CUTLASS          & 49.96       & 0.67 \\ \midrule
Native\_kernel   & 20.05       & 1.67 \\
+ Pipeline Optimization    & 14.66       & 2.28 \\
+ Eliminiate GEMV & 10.92       & 3.07 \\
+ Auto Kernel Search     & \textbf{6.68} & \textbf{5.01} \\ \bottomrule
\end{tabular}
\caption{An ablation study of the impact of optimization techniques used in the inference engine on Kernel latency and throughput.}
\label{tab:kernel_ablation}
\end{table}

\section{Conclusion}
We present an arbitrary bit quantization inference framework called ABQ-LLM. Through an in-depth analysis of LLM quantization, we introduce distribution correction and bit balance strategy to enhance model performance. We then design a novel arbitrary bit inference engine to fully leverage the advantages of LLM quantization. Extensive experimental results demonstrate that ABQ-LLM achieves outstanding performance across various quantization configurations, including W6A6, W4A4, and W2A8. Moreover, ABQKernel consistently outperformed both CUTLASS and cuBLAS in all configurations. Our end-to-end inference speed is 1.6 $\times$ faster than the industry SOTA, SmoothQuant, and achieves 2.7 $\times$ memory compression gain.


\bibliography{aaai25}

\newpage
\appendix
\onecolumn

\section{Bit Balance Analyze} \label{a:bit_analyse}
\begin{figure*}[ht]
\centering
\includegraphics[width=0.98\textwidth]{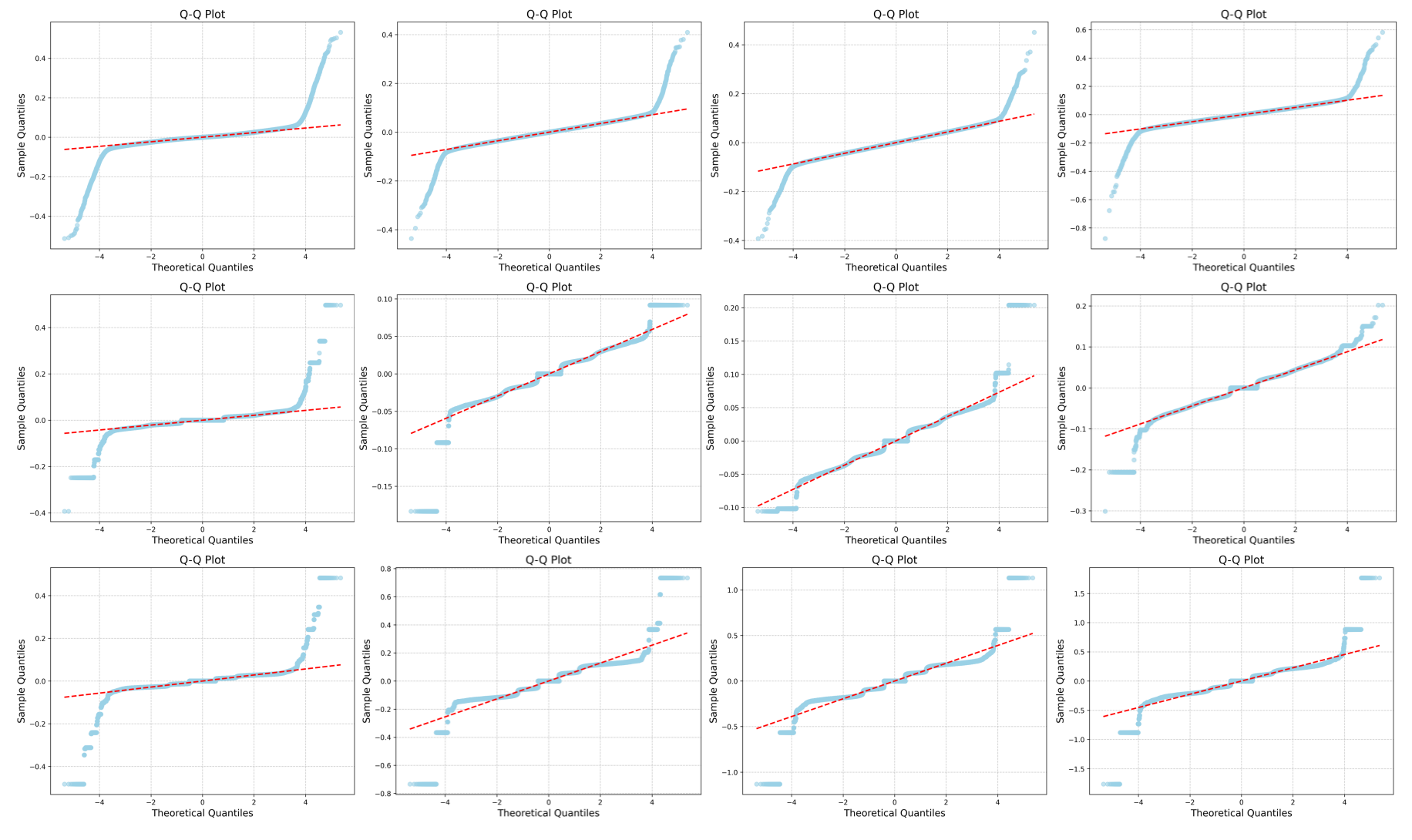}
\caption{Quantile-Quantile Plots of data distribution with o\_proj weights in the 1st, 10th, 20th, and 30th blocks at full precision (up), INT2 quantization (middle) and INT2* quantization (below) in the LLaMA-7B model}
\label{fig:qq}
\end{figure*}

As shown in Figure~\ref{fig:qq}, there is a clear asymmetry in the weights of o\_proj after INT2 quantization, particularly evident in the 10th and 20th blocks. This asymmetry shifts the distribution of the model's weights, resulting in significant performance loss. However, after applying the balancing strategy, the distribution of weights closely aligns with that of the full-precision model, as illustrated in the subsequent figure. This alignment preserves the original symmetric distribution of weights and greatly enhances model performance.

\section{Arbitrary Bit Matrix Calculation} \label{a:abc}
Formally, given a p-bit weight matrix W and a q-bit activation matrix X, we can first decompose into 1-bit matrices $W^{s}, \quad s \in \left \{ 0,1,...,p-1 \right \}$ and $X^{t}, \quad t \in \left \{ 0,1,...,p-1 \right \}$.
\begin{equation}
\begin{aligned} 
w^{s}_{i,j} = \left( w_{i,j} \gg s \right) \& 1, \quad x^{t}_{i,j} = \left( x_{i,j} \gg t \right) \& 1
\end{aligned}
\end{equation}
If BMMA stands for 1-bit matrix multiplication, then the BMMA operation can be called $p \times q$ times to compute a series of 1-bit matrix multiplication components
\begin{equation}
\begin{aligned} 
Y^{s,t} = BMMA\left( W^{s}, X^{t} \right) 
\end{aligned}
\end{equation}
Finally, we process all 1-bit matrix multiplication components with bit-stacked scaling factors, after which they are summed to obtain a 32-bit output matrix
\begin{equation}
\begin{aligned} 
Y = \sum_{s=0}^{p-1}\sum_{t=0}^{q-1} Y^{s,t}\ast 2^{s+t} 
\end{aligned}
\label{formula:matrix}
\end{equation}
After the above transformation process, we decompose the operation of arbitrary quantized combinations into a special superposition of 1-bit matrix multiplications, so that the underlying layer can be implemented by calling high-computing BMMA instructions.

\section{Full Results for Quantization Algorithms} \label{quantization_all_result}

In this paper, we mainly publish metrics for per-channel/per-token quantization. However, ABQ-LLM is naturally orthogonal to per-group quantization, and we validate this on LLaMA-7B for W4A4 g128 per-group quantization with the results shown in Table~\ref{group-quant-ppl}. Experimental results demonstrate that, in most cases, our method outperforms another state-of-the-art (SOTA) method, Atom\cite{zhao2024atom}, which employs per-group quantization. Additionally, our method exhibits a degradation of less than 0.5 in PPL compared to FP16, indicating excellent performance.

Furthermore, we present the perplexity (PPL) metrics and zero-shot performance of ABQ-LLM under various quantization configurations. Unfortunately, most prior work has not disclosed complete experimental data, with only PPL metrics for common configurations such as W4A16, W3A16, W2A16, W4A4, and W6A6 being published. Additionally, the full zero-shot results were not made publicly available. We aim to establish a comprehensive quantization baseline for post-training quantization (PTQ) by making our extensive data publicly accessible. However, due to limitations in time and computational resources, we can only provide our own comprehensive experimental results and cannot fully reproduce the results of other baselines.
\begin{table}[!h]
\centering
\begin{tabular}{ccccccccccc}
\toprule
\multirow{2}{*}{Bits} & \multirow{2}{*}{Method} & \multicolumn{2}{c}{PPL$\downarrow$} & \multicolumn{7}{c}{Accuracy(\%)$\uparrow$} \\ \cmidrule(lr){3-4} \cmidrule(l){5-11} 
 &  & Wikitext2 & C4 & \multicolumn{1}{l}{PiQA} & \multicolumn{1}{l}{ARC-e} & \multicolumn{1}{l}{ARC-c} & \multicolumn{1}{l}{BoolQ} & \multicolumn{1}{l}{HellaSwag} & \multicolumn{1}{l}{Winogrande} & Avg. \\ \midrule
FP16 & - & 5.67 & 7.08 & 77.47 & 52.48 & 41.46 & 73.08 & 73.00 & 67.07 & 64.09 \\
W4A4 g128 & Atom & 6.16 & 7.70 & 76.28 & \textbf{52.10} & \textbf{38.99} & 69.79 & 69.81 & \textbf{63.69} & 61.78 \\
W4A4 g128 & ABQ-LLM & \textbf{6.05} & \textbf{7.61} & \textbf{76.55} & 51.81 & 38.14 & \textbf{71.65} & \textbf{70.22} & 63.22 & \textbf{61.93} \\ \bottomrule
\end{tabular}
\caption{Performance of weight-activation per-group quantization on LLaMA-7B with a group size of 128.}
\label{group-quant-ppl}
\end{table}

To ensure the reproducibility of our results, we specify that for the PPL test, we follow the GPTQ method by setting the sentence length to 2048. It is important to note that different sentence lengths may affect the accuracy of the metrics. Additionally, for the zero-shot accuracy test, two metrics may be used: \textit{acc} or \textit{acc\_norm}. If \textit{acc\_norm} is available, we report \textit{acc\_norm}; otherwise, we report the \textit{acc} metric. Finally, we have published the quantized model series on the Huggingface platform. Table~\ref{weight-only} presents the perplexity metrics for weight-only quantization, while Table~\ref{tab:other-setting} displays the perplexity metrics for various configurations where both weights and activations are quantized simultaneously. The zero-shot accuracies of LLaMA-7B, LLaMA-13B, LLaMA-2-7B, and LLaMA-2-13B are shown in Tables~\ref{tab:llama-7b -zero-shot}, \ref{tab:llama-13b -zero-shot}, \ref{tab:llama-2-7b -zero-shot}, and \ref{tab:llama-2-13b -zero-shot}, respectively. Extensive experiments demonstrate that our method consistently outperforms other baselines across all quantization configurations, achieving state-of-the-art performance in both weight-only quantization and weight-activation quantization.

\begin{table*}[ht]
\centering
\begin{tabular}{cccccccccc}
\toprule
\multirow{2}{*}{Bits} &
  \multirow{2}{*}{Method} &
  \multicolumn{2}{c}{LLaMA-7B} &
  \multicolumn{2}{c}{LLaMA-13B} &
  \multicolumn{2}{c}{LLaMA-2-7B} &
  \multicolumn{2}{c}{LLaMA-2-13B} \\ \cmidrule(l){3-10} 
                       &                  & WikiText2     & C4            & WikiText2     & C4            & WikiText2     & C4             & WikiText2     & C4            \\ \midrule
FP16                   & -                & 5.67          & 7.08          & 5.09          & 6.61          & 5.47          & 6.97           & 4.88          & 6.46          \\ \midrule
\multirow{5}{*}{W4A16} & GPTQ             & 6.13          & 7.43          & 5.40          & 6.84          & 5.83          & 7.37           & 5.13          & 6.70          \\
                       & AWQ              & 6.08          & 7.52          & 5.34          & 6.86          & 6.15          & 7.68           & 5.12          & 6.74          \\
                       & OmniQuant        & 5.86          & 7.34          & 5.21          & 6.76          & 5.74          & 7.35           & 5.02          & 6.65          \\
                       & AffineQuant      & 5.84          & 7.30          & 5.20          & 6.75          & 5.69          & 7.29           & 5.01          & 6.64          \\
                       & \textbf{ABQ-LLM} & \textbf{5.83} & \textbf{7.29} & \textbf{5.19} & \textbf{6.75} & \textbf{5.64} & \textbf{7.20}  & \textbf{5.01} & \textbf{6.63} \\ \midrule
\multirow{5}{*}{W3A16} & GPTQ             & 8.06          & 9.49          & 6.76          & 8.16          & 8.37          & 9.81           & 6.44          & 8.02          \\
                       & AWQ              & 11.88         & 13.26         & 7.45          & 9.13          & 24.00         & 23.85          & 10.45         & 13.07         \\
                       & OmniQuant        & 6.49          & 8.19          & 5.68          & 7.32          & 6.58          & 8.65           & 5.58          & 7.44          \\
                       & AffineQuant      & 6.30          & 8.03          & 5.60          & 7.20          & 6.55          & 8.57           & 5.62          & 7.56          \\
                       & \textbf{ABQ-LLM} & \textbf{6.29} & \textbf{8.01} & \textbf{5.56} & 7.24 & \textbf{6.28} & \textbf{8.10}  & \textbf{5.44} & \textbf{7.26} \\ \midrule
\multirow{4}{*}{W2A16} & GPTQ             & 2.1e3         & 689.13        & 5.5e3         & 2.5e3         & 7.7e3         & NAN            & 2.1e3         & 323.12        \\
                       & OmniQuant        & 15.47         & 24.89         & 13.21         & 18.31         & 37.37         & 90.64          & 17.21         & 26.76         \\
                       & AffineQuant      & \textbf{9.53}  & \textbf{14.89}         & \textbf{7.54}          &       12.46       & 35.07         &  572.22    & \textbf{12.42} &   23.67  \\
                       & \textbf{ABQ-LLM} & 11.48 & 15.74 & 9.34 & \textbf{12.28} & \textbf{13.11} & \textbf{17.81} & 13.09 & \textbf{20.49} \\ \midrule
                W2*A16       & \textbf{ABQ-LLM} & \textbf{7.50} & \textbf{9.86} & \textbf{6.64} & \textbf{8.43} & \textbf{7.82} & \textbf{10.33} & \textbf{6.52} & \textbf{7.87} \\ \bottomrule
\end{tabular}
\caption{Weight-only quantization perplexities (lower is better) comparison of quantized LLaMA and LLaMA-2 models. * denotes the use of the bit balance strategy.}
\label{weight-only}
\end{table*}

\begin{table*}[!ht]
\centering
\begin{tabular}{cccccccccc}
\toprule
\multirow{2}{*}{Bits} & \multirow{2}{*}{Method} & \multicolumn{2}{c}{LLaMA-7B} & \multicolumn{2}{c}{LLaMA-13B} & \multicolumn{2}{c}{LLaMA-2-7B} & \multicolumn{2}{c}{LLaMA-2-13B} \\ \cmidrule(l){3-10} 
 &  & Wikitext2 & C4 & Wikitext2 & C4 & Wikitext2 & C4 & Wikitext2 & C4 \\ \midrule
FP16 & - & 5.67 & 7.08 & 5.09 & 6.61 & 5.47 & 6.97 & 4.88 & 6.46 \\ \midrule
\multirow{2}{*}{W8A8} & SmoothQuant & 5.73 & - & 5.13 & - & 5.54 & - & 4.95 & - \\
 & ABQ-LLM & \textbf{5.68} & \textbf{7.09} & \textbf{5.10} & \textbf{6.62} & \textbf{5.48} & \textbf{6.99} & \textbf{4.89} & \textbf{6.47} \\ \midrule
\multirow{6}{*}{W4A8} & AWQ & 6.33 & - & 5.59 & - & 6.28 & - & 5.25 & - \\
 & QuaRot & 5.93 & - & 5.29 & - & 5.73 & - & 5.07 & - \\
 & Atom & 6.03 & - & 5.41 & - & 5.91 & - & 5.16 & - \\
 & Qserve & 5.93 & - & 5.28 & - & 5.75 & - & 5.12 & - \\
 & OmniQuant & 5.87 & 7.34 & - & - & - & - & - & - \\
 & ABQ-LLM & \textbf{5.84} & \textbf{7.32} & \textbf{5.22} & \textbf{6.77} & \textbf{5.67} & \textbf{7.24} & \textbf{5.01} & \textbf{6.64} \\ \midrule
\multirow{2}{*}{W4A6} & OmniQuant & 6.09 & 7.63 & - & - & - & - & - & - \\
 & ABQ-LLM & \textbf{6.01} & \textbf{7.58} & 5.37 & 6.96 & 5.89 & 7.56 & 5.17 & 6.87 \\ \midrule
W3A8 & ABQ-LLM & 6.30 & 8.04 & 5.59 & 7.26 & 6.27 & 8.14 & 5.45 & 7.27 \\ \midrule
W3A6 & ABQ-LLM & 6.60 & 8.47 & 5.84 & 7.61 & 6.56 & 8.65 & 5.92 & 7.90 \\ \midrule
W3A4 & ABQ-LLM & 12.16 & 17.19 & 9.96 & 14.36 & 13.65 & 19.00 & 20.35 & 20.09 \\ \midrule
W2*A8 & ABQ-LLM & 7.59 & 10.00 & 6.49 & 8.53 & 7.85 & 10.33 & 6.65 & 10.01 \\ \midrule
W2*A6 & ABQ-LLM & 8.08 & 10.89 & 6.99 & 9.30 & 9.08 & 11.74 & 10.91 & 14.89 \\ \bottomrule
\end{tabular}
\caption{Weight-activation quantization perplexities (lower is better) comparison of quantized LLaMA and LLaMA-2 models. * denotes the use of the bit balance strategy.}
\label{tab:other-setting}
\end{table*}

\begin{table*}[!ht]
\centering
\begin{tabular}{cccccccccc}
\toprule
Model & Bits & Method & PiQA & ARC-e & ARC-c & BoolQ & HellaSwag & Winogrande & Avg. \\ \midrule
\multirow{14}{*}{LLaMA-2-7B} & FP16 & - & 76.98 & 53.57 & 40.61  & 71.07  &  72.96 &  67.24 & 63.73 \\ \cmidrule(l){2-10} 
 & W4A16 & ABQ-LLM & 77.36 & 53.82 & 39.50 & 70.76 & 71.76 & 66.85 & 63.34 \\ \cmidrule(l){2-10} 
 & W3A16 & ABQ-LLM & 76.55 & 53.66 & 39.24 & 63.97 & 68.93 & 65.90 & 61.38 \\ \cmidrule(l){2-10} 
 & W2*A16 & ABQ-LLM & 72.79 & 48.14 & 35.32 & 63.94 & 63.21 & 61.79 & 57.53 \\ \cmidrule(l){2-10} 
 & W8A8 & ABQ-LLM & 76.76 & 53.53 & 40.35 & 71.19 & 72.84 & 66.92 & 63.60 \\ \cmidrule(l){2-10} 
 & W6A6 & ABQ-LLM & 76.77 & 53.20 & 40.44 & 71.37 & 71.81 & 66.22 & 63.30 \\ \cmidrule(l){2-10} 
 & W4A8 & ABQ-LLM & 76.76 & 53.11 & 39.07 & 67.86 & 71.38 & 66.92 & 62.52 \\ \cmidrule(l){2-10} 
 & W4A6 & ABQ-LLM & 76.06 & 52.56 & 38.82 & 67.22 & 70.35 & 64.33 & 61.60 \\ \cmidrule(l){2-10} 
 & W4A4 & ABQ-LLM & 68.55 & 44.11 & 31.31 & 63.12 & 55.29 & 53.82 & 52.70 \\ \cmidrule(l){2-10} 
 & W3A8 & ABQ-LLM & 76.55 & 52.44 & 38.99 & 65.77 & 68.34 & 66.61 & 61.45 \\ \cmidrule(l){2-10} 
 & W3A6 & ABQ-LLM & 74.26 & 51.47 & 37.62 & 65.22 & 66.74 & 63.22 & 59.76 \\ \cmidrule(l){2-10} 
 & W3A4 & ABQ-LLM & 64.47 & 39.56 & 27.47 & 58.81 & 49.28 & 53.28 & 48.81 \\ \cmidrule(l){2-10} 
 & W2*A8 & ABQ-LLM & 72.74 & 47.81 & 35.07 & 62.78 & 63.31 & 61.09 & 57.13 \\ \cmidrule(l){2-10} 
 & W2*A6 & ABQ-LLM & 71.98 & 45.54 & 31.66 & 62.14 & 60.14 & 55.25 & 54.45 \\ \bottomrule 
\end{tabular}
\caption{ Zero-shot accuracy (higher is better) of LLaMA-2-7B under different quantization configuration. * denotes the use of the bit balance strategy.}
\label{tab:llama-2-7b -zero-shot}
\end{table*}

\begin{table*}[!ht]
\centering
\begin{tabular}{cccccccccc}
\toprule
Model & Bits & Method & PiQA & ARC-e & ARC-c & BoolQ & HellaSwag & Winogrande & Avg. \\ \midrule
\multirow{21}{*}{LLaMA-7B} & FP16 & - & 77.47 & 52.48 & 41.46 & 73.08 & 73.00 & 67.07 & 64.09 \\ \cmidrule(l){2-10} 
 & \multirow{2}{*}{W4A16} & AffineQuant & 77.53 & \textbf{51.85} & 38.65 & 70.89 & \textbf{71.49} & \textbf{66.93} & 62.89 \\
 &  & ABQ-LLM & \textbf{77.80} & 51.55 & \textbf{39.59} & \textbf{72.66} & 71.41 & 66.14 & \textbf{63.19} \\ \cmidrule(l){2-10} 
 & W3A16 & ABQ-LLM & 75.78 & 47.81 & 37.20 & 71.98 & 68.80 & 65.82 & 61.23 \\ \cmidrule(l){2-10} 
 & W2*A16 & ABQ-LLM & 74.04 & 46.71 & 35.49 & 65.81 & 48.01 & 62.82 & 55.48 \\ \cmidrule(l){2-10} 
 & W8A8 & ABQ-LLM & 77.25 & 52.56 & 41.72 & 72.97 & 73.05 & 67.32 & 64.14 \\ \cmidrule(l){2-10} 
 & \multirow{4}{*}{W6A6} & SmoothQuant & 76.75 & 51.64 & 39.88 & 71.75 & 71.67 & 65.03 & 62.81 \\
 &  & OmniQuant & 77.09 & 51.89 & \textbf{40.87} & 72.53 & 71.61 & 65.03 & 63.17 \\
 &  & I-LLM & 76.99 & 52.66 & 40.78 & \textbf{72.94} & 71.31 & 65.67 & 63.39 \\
 &  & ABQ-LLM & \textbf{78.07} & \textbf{52.81} & 40.10 & 71.90 & \textbf{71.80} & \textbf{66.53} & \textbf{63.53} \\ \cmidrule(l){2-10} 
 & \multirow{2}{*}{W4A8} & OmniQuant & 77.36 & 51.85 & \textbf{38.65} & 70.67 & 71.20 & 64.71 & 62.40 \\
 &  & ABQ-LLM & \textbf{77.63} & \textbf{52.40} & 38.10 & \textbf{73.43} & \textbf{71.81} & \textbf{65.51} & \textbf{63.14} \\ \cmidrule(l){2-10} 
 & \multirow{2}{*}{W4A6} & OmniQuant & 75.73 & \textbf{51.52} & 38.31 & 68.28 & \textbf{70.79} & \textbf{65.27} & 61.64 \\
 &  & ABQ-LLM & \textbf{76.28} & 50.72 & \textbf{39.68} & \textbf{70.52} & 70.35 & 64.88 & \textbf{62.06} \\ \cmidrule(l){2-10} 
 & \multirow{3}{*}{W4A4} & OmniQuant & 66.15 & 45.20 & 31.14 & 63.51 & 56.44 & 53.43 & 52.65 \\
 &  & AffineQuant & 69.37 & 42.55 & 31.91 & \textbf{63.73} & 57.65 & \textbf{55.33} & 53.42 \\
 &  & ABQ-LLM & \textbf{69.97} & \textbf{45.88} & \textbf{33.44} & 62.87 & \textbf{58.47} & 54.53 & \textbf{54.19} \\ \cmidrule(l){2-10} 
 & W3A8 & ABQ-LLM & 75.78 & 48.95 & 38.05 & 70.18 & 68.70 & 65.74 & 61.23 \\ \cmidrule(l){2-10} 
 & W3A6 & ABQ-LLM & 75.57 & 49.03 & 38.57 & 69.57 & 67.12 & 62.35 & 60.37 \\ \cmidrule(l){2-10} 
 & W3A4 & ABQ-LLM & 64.20 & 41.04 & 28.50 & 61.80 & 49.71 & 51.54 & 49.45 \\ \cmidrule(l){2-10} 
 & W2*A8 & ABQ-LLM & 73.29 & 46.72 & 34.73 & 63.61 & 62.18 & 61.01 & 56.92 \\ \cmidrule(l){2-10} 
 & W2*A6 & ABQ-LLM & 71.98 & 46.00 & 33.53 & 63.95 & 59.19 & 58.72 & 55.56 \\ \bottomrule
\end{tabular}
\caption{ Zero-shot accuracy (higher is better) of LLaMA-7B under different quantization configuration. * denotes the use of the bit balance strategy.}
\label{tab:llama-7b -zero-shot}
\end{table*}

\begin{table*}[!ht]
\centering
\begin{tabular}{lccccccccc}
\toprule
Model & \multicolumn{1}{l}{Bits} & \multicolumn{1}{l}{Method} & \multicolumn{1}{l}{PiQA} & \multicolumn{1}{l}{ARC-e} & \multicolumn{1}{l}{ARC-c} & \multicolumn{1}{l}{BoolQ} & \multicolumn{1}{l}{HellaSwag} & \multicolumn{1}{l}{Winogrande} & Avg. \\ \midrule
\multirow{21}{*}{LLaMA-13B} & FP16 & - & 79.10 & 59.89 & 44.45 & 68.01 & 76.21 & 70.31 & 66.32 \\ \cmidrule(l){2-10} 
 & \multirow{2}{*}{W4A16} & AffineQuant & 78.84 & \textbf{59.55} & \textbf{43.52} & 69.48 & 75.18 & \textbf{69.38} & \multicolumn{1}{c}{65.99} \\
 &  & ABQ-LLM & \textbf{78.89} & 59.43 & 43.25 & \textbf{70.37} & \textbf{75.38} & 69.22 & \textbf{66.09} \\ \cmidrule(l){2-10} 
 & W3A16 & ABQ-LLM & 77.86 & 57.62 & 42.15 & 66.45 & 73.03 & 69.53 & 64.44 \\ \cmidrule(l){2-10} 
 & W2*A16 & ABQ-LLM & 75.03 & 52.99 & 38.48 & 66.42 & 68.69 & 66.53 & 61.36 \\ \cmidrule(l){2-10} 
 & W8A8 & ABQ-LLM & 78.73 & 59.01 & 44.11 & 68.17 & 75.96 & 70.09 & 66.01 \\ \cmidrule(l){2-10} 
 & \multirow{4}{*}{W6A6} & SmoothQuant & 77.91 & 56.60 & 42.40 & 64.95 & 75.36 & 69.36 & \multicolumn{1}{c}{64.43} \\
 &  & OmniQuant & 78.40 & 57.28 & 42.91 & \textbf{67.00} & \textbf{75.82} & 68.27 & \multicolumn{1}{c}{64.94} \\
 &  & I-LLM & 77.48 & 56.94 & \textbf{44.03} & 64.92 & 75.24 & 69.14 & \multicolumn{1}{c}{64.62} \\
 &  & ABQ-LLM & \textbf{78.40} & \textbf{57.62} & 42.01 & 66.82 & 75.54 & \textbf{69.61} & \multicolumn{1}{c}{\textbf{65.00}} \\ \cmidrule(l){2-10} 
 & W4A8 & ABQ-LLM & 78.56 & 58.75 & 42.92 & 68.56 & 75.20 & 70.56 & 65.76 \\ \cmidrule(l){2-10} 
 & W4A6 & ABQ-LLM & 77.64 & 56.94 & 42.32 & 67.06 & 74.23 & 68.19 & 64.39 \\ \cmidrule(l){2-10} 
 & \multirow{4}{*}{W4A4} & OmniQuant & 69.69 & 47.30 & 33.10 & 62.84 & 58.96 & 55.80 & \multicolumn{1}{c}{54.62} \\
 &  & AffineQuant & 66.32 & 43.90 & 29.61 & \textbf{64.10} & 56.88 & 54.70 & \multicolumn{1}{c}{52.58} \\
 &  & I-LLM & 67.95 & \textbf{48.15} & 34.47 & 62.29 & 63.13 & \textbf{59.98} & \multicolumn{1}{c}{55.99} \\
 &  & ABQ-LLM & \textbf{71.82} & 47.60 & \textbf{35.67} & 63.52 & \textbf{64.31} & 57.54 & \multicolumn{1}{c}{\textbf{56.74}} \\ \cmidrule(l){2-10} 
 & W3A8 & ABQ-LLM & 77.81 & 58.16 & 42.41 & 68.47 & 73.15 & 69.37 & \multicolumn{1}{c}{64.90} \\ \cmidrule(l){2-10} 
 & W3A4 & ABQ-LLM & 64.79 & 42.21 & 30.54 & 60.55 & 55.59 & 53.51 & 51.20 \\ \cmidrule(l){2-10} 
 & W3A6 & ABQ-LLM & 76.71 & 56.19 & 40.53 & 66.39 & 71.59 & 66.14 & 62.93 \\ \cmidrule(l){2-10} 
 & W2*A8 & ABQ-LLM & 74.92 & 54.92 & 38.65 & 68.53 & 68.21 & 66.54 & 61.96 \\ \cmidrule(l){2-10} 
 & W2*A6 & ABQ-LLM & 73.71 & 51.89 & 36.60 & 64.98 & 65.55 & 63.22 & 59.33 \\ \bottomrule
\end{tabular}
\caption{ Zero-shot accuracy (higher is better) of LLaMA-13B under different quantization configuration. * denotes the use of the bit balance strategy.}
\label{tab:llama-13b -zero-shot}
\end{table*}

\begin{table*}[!ht]
\centering
\begin{tabular}{cccccccccc}
\toprule
\multicolumn{1}{c}{Model} & \multicolumn{1}{c}{Bits} & \multicolumn{1}{c}{Method} & \multicolumn{1}{c}{PiQA} & \multicolumn{1}{c}{ARC-e} & \multicolumn{1}{c}{ARC-c} & \multicolumn{1}{c}{BoolQ} & \multicolumn{1}{c}{HellaSwag} & \multicolumn{1}{c}{Winogrande} & Avg. \\ \midrule
\multirow{13}{*}{LLaMA-2-13B} & FP16 & - & 79.05 & 57.91 & 44.19 & 69.02 & 76.60 & 69.69 & \multicolumn{1}{c}{66.07} \\ \cmidrule(l){2-10} 
 & W4A16 & ABQ-LLM & 78.72 & 58.16 & 44.03 & 63.94 & 75.56 & 69.13 & 64.92 \\ \cmidrule(l){2-10} 
 & W3A16 & ABQ-LLM & 77.48 & 55.39 & 43.86 & 67.58 & 72.63 & 67.56 & 64.08 \\ \cmidrule(l){2-10} 
 & W2*A16 & ABQ-LLM & 75.41 & 51.39 & 36.43 & 72.63 & 67.03 & 60.62 & 60.59 \\ \cmidrule(l){2-10} 
 & W8A8 & ABQ-LLM & 79.22 & 57.66 & 43.94 & 68.62 & 76.47 & 69.46 & 65.90 \\ \cmidrule(l){2-10} 
 & W6A6 & ABQ-LLM & 78.62 & 55.81 & 43.69 & 66.33 & 75.51 & 68.27 & 64.70 \\ \cmidrule(l){2-10} 
 & W4A8 & ABQ-LLM & 78.67 & 57.15 & 43.52 & 64.13 & 75.31 & 69.53 & 64.53 \\ \cmidrule(l){2-10} 
 & W4A6 & ABQ-LLM & 77.97 & 57.11 & 43.26 & 67.37 & 74.14 & 66.30 & 64.36 \\ \cmidrule(l){2-10} 
 & W4A4 & ABQ-LLM & 69.04 & 47.01 & 33.53 & 64.74 & 62.70 & 54.38 & 55.23 \\ \cmidrule(l){2-10} 
 & W3A8 & ABQ-LLM & 77.53 & 56.36 & 42.74 & 68.71 & 72.87 & 66.06 & 64.05 \\ \cmidrule(l){2-10} 
 & W3A6 & ABQ-LLM & 76.22 & 53.37 & 40.70 & 68.04 & 71.09 & 66.14 & 62.59 \\ \cmidrule(l){2-10} 
 & W3A4 & ABQ-LLM & 63.60 & 42.55 & 29.35 & 58.62 & 52.62 & 53.19 & 49.99 \\ \cmidrule(l){2-10} 
 & W2*A8 & ABQ-LLM & 73.50 & 49.79 & 35.15 & 70.15 & 67.45 & 58.88 & 59.15 \\ \bottomrule
\end{tabular}
\caption{ Zero-shot accuracy (higher is better) of LLaMA-2-13B under different quantization configuration. * denotes the use of the bit balance strategy.}
\label{tab:llama-2-13b -zero-shot}
\end{table*}

\clearpage
\section{GPU Kernel Optimization Details} 
\label{gpu_opt}
\begin{figure}[!htb]
\centering
\includegraphics[width=\linewidth]{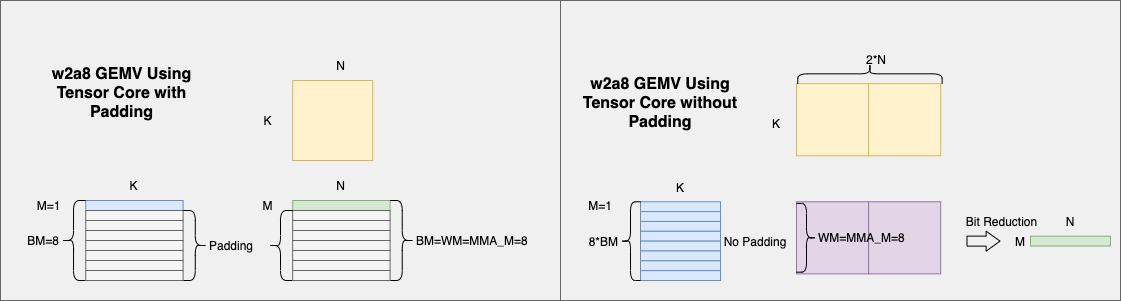}
\caption{Overview of GEMV Elimination. }
\label{fig:gemv_elim}
\end{figure}

\noindent \textbf{GEMV Elimination.} Our custom engine decomposes the operation of arbitrary quantized combinations into a superposition of 1-bit matrix multiplications. As for GEMV problems, multiple single-bit GEMV are converted to GEMM, fully utilizing the TensorCore while reducing or even avoiding the redundant computation caused by padding. For the $W_{q}A_{p}$ configuration, the actual computing task undertaken by ABQKernel is p*M × q*N × K. When p*M \% MMA\_M = 0, we can achieve zero redundant calculation. As shown in Figure \ref{fig:gemv_elim}, there are two problems with directly calling INT8 TensorCore to execute W2A8 GEMV: 1. Converting W2 to W8 online requires additional instructions. 2. When M $=$ 1 and MMA\_M $=$ 8, calling TensorCore will introduce 87.5\% redundant calculations. In contrast, ABQKernel achieves direct acceleration of W2A8 without any redundant computation. 

\begin{figure}[!htb]
\centering
\includegraphics[width=\linewidth]{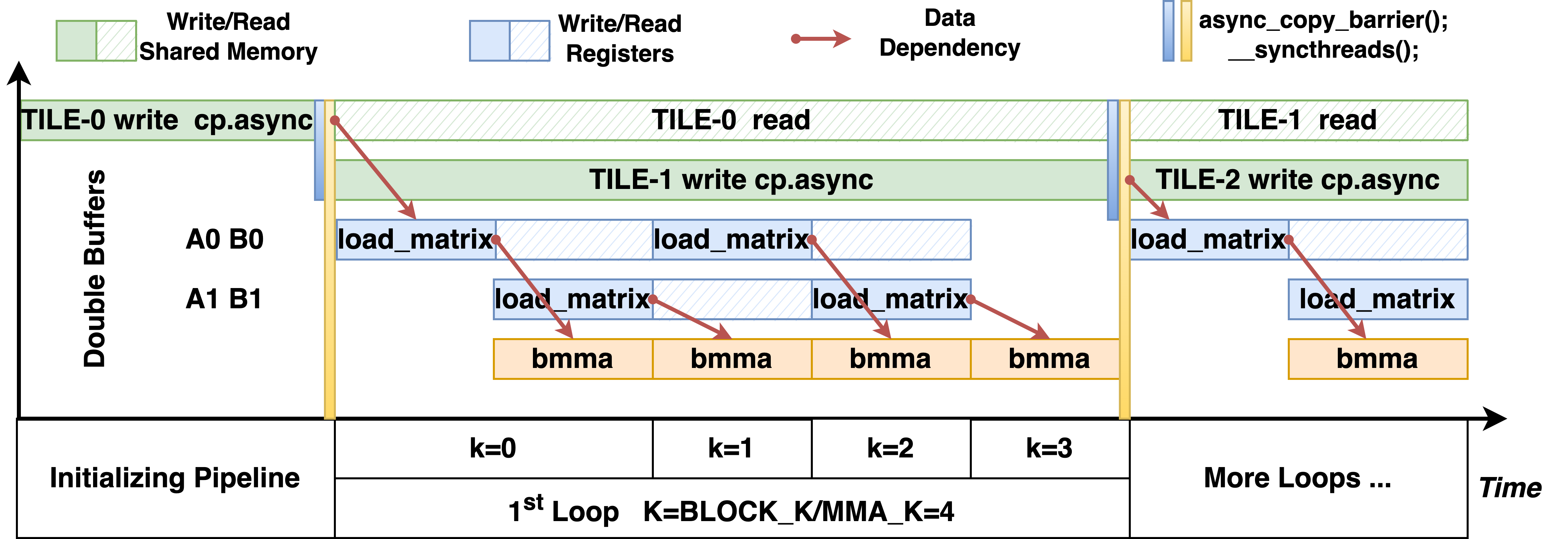}
\caption{Space-time diagram of the computational pipeline.}
\label{fig:double_buffer}
\end{figure}

\noindent \textbf{Computational Pipeline Optimization.} Figure~\ref{fig:double_buffer} illustrates our optimization of the computation pipeline to enhance inference performance within the widely adopted Ampere architecture. At the shared memory level, we perform asynchronous writes from global memory to shared memory using the \textit{cp.async} instruction to hide the latency of subsequent memory accesses. Before processing the first loop, a synchronization instruction ensures that TILE-0, the data required by the first loop, has been written to shared memory. During the processing of the first loop, TILE-1, the data needed for the second loop, is asynchronously written to shared memory. This concurrent execution of data writes for the second loop and computation of the first loop masks the shared memory access latency for subsequent loops. For register-level optimization within the loop, when k=0, TILE-0 data from shared memory is loaded into the first set of registers \textit{A0} and \textit{B0}, while the data required for k=1 is preloaded into registers \textit{A1} and \textit{B1}. Once the data in registers \textit{A0} and \textit{B0} is ready, the bmma operation is performed on \textit{A0} and \textit{B0}. When k=1, the data \textit{A1} and \textit{B1} needed for bmma has already been preloaded at k=0, and the data required for k=2 is preloaded into registers \textit{A0} and \textit{B0}, and so on. By doubling the register cache, data is written from shared memory to registers during the Tensor Core bmma computation, effectively masking the register access time. Where bmma denotes a 1-bit matrix multiplication operation, while TILE data refers to a block of data obtained after partitioning and processing the original dataset.

\noindent \textbf{Auto Kernel Search.} To utilize GPU memory more efficiently at all levels, a chunking strategy is essential during computation. Chunk size is crucial for GEMM performance. Before launching an arbitrary precision inference operator, we perform performance tests on different chunk sizes to select the best implementation. As shown in Figure~\ref{fig:mmba}, for the classical GEMM task, the M × N × K problem is usually chunked by BM × BN × BK at the thread block level, and each thread block is chunked by WM × WN × WK at the warp level. Each warp gets MMA\_M × MMA\_N × MMA\_K based on the Tensor Core chunk size supported by the GPU model and searches for different chunk shapes after obtaining the chunk sizes at the three levels. In the arbitrary precision operator, the number of weight bits (q) and activation bits (p) also need to be considered, so the search space is larger compared to the classical GEMM. Given the BTC chunk sizes MMA\_M = 8, MMA\_N = 8, MMA\_K = 128, the number of weighted warps is computed as W\_WARPS\_NUM = BN × q / WN, and the number of activation warps is computed as X\_WARPS\_NUM = BM × p / WM. The total number of warps is 1 $\le$ X\_WARPS\_NUM × W\_WARPS\_NUM $\le$ 32. To reduce the search space, we fix WK = MMA\_K = 128 and set the length of BK to {128, 256, 384, 512}. BM, BN, BK, WM, and WN cannot be infinitely large due to the limited shared memory and register usage of a single thread block. As for $W_{q}A_{p}$ configuration, the design process of a candidate instance in the search space is summarized as follows: 

1. Determine the number and layout of Warps contained in the Thread Block based on expert experience. for example, X\_WARPS\_NUM × W\_WARPS\_NUM = 1 × 4.

2. Determine the Thread Block Tile($<$BM, BN, BK$>$) with the smallest redundant padding based on the quantized bit width p of the activation and the M dimension of the computing task. 

3. The size of the WARP Tile($<$WM, WN, WK$>$) is calculated based on the layout of Warps and Thread Block Tile. Ultimately, we test the operators at various chunk sizes and adopt the speed-optimized implementation.

\begin{figure}[!htb]
\centering
\includegraphics[width=\linewidth]{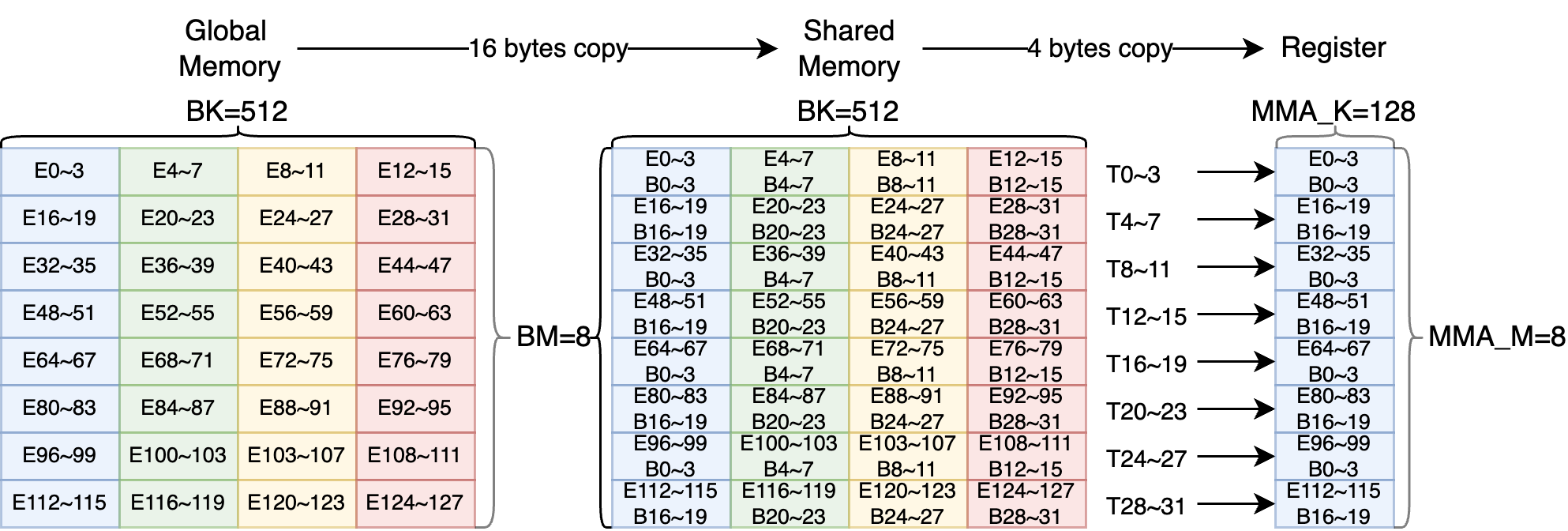}
\caption{Address occurs shared memory 4-way bank conflicts. E denotes an \textbf{e}lement of int32, B denotes \textbf{b}ank, and T denotes a \textbf{t}hread.}
\label{fig:bank-conflict}
\end{figure}

\begin{figure}[!htb]
\centering
\includegraphics[width=\linewidth]{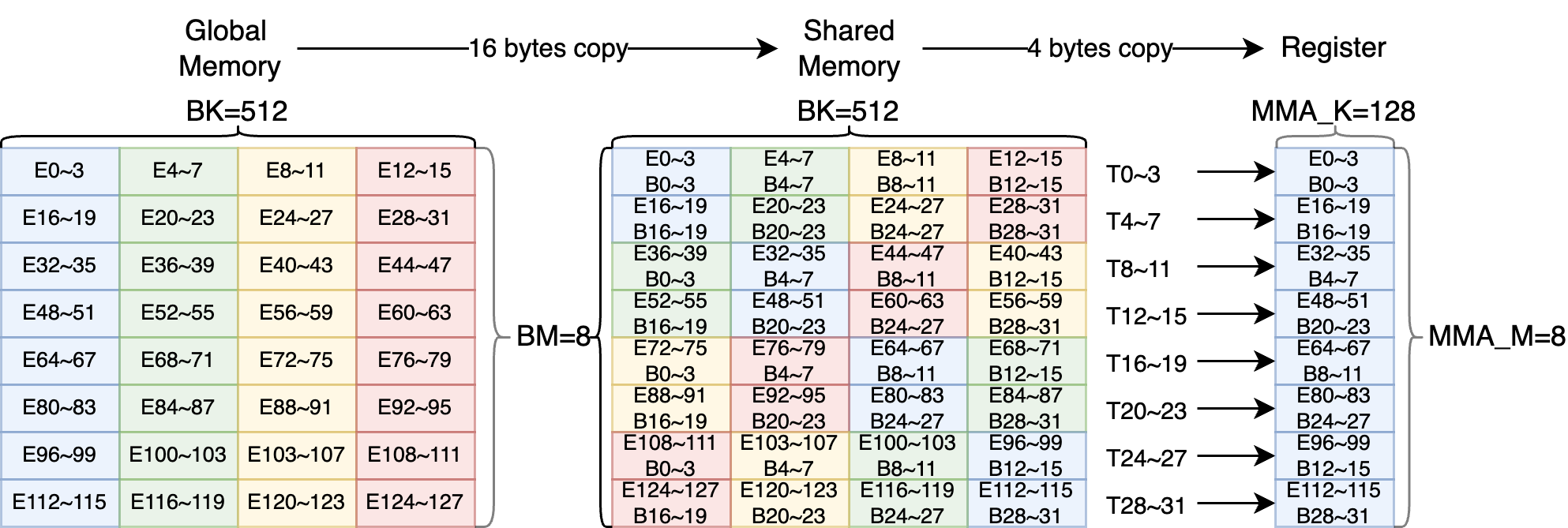}
\caption{Address remapping avoids shared memory bank conflicts. E denotes an element of int32, B denotes bank, and T denotes
a thread.}
\label{fig:bank-conflict-solution}
\end{figure}

\noindent \textbf{Bank Conflicts Elimination.} For attaining high bandwidth, shared memory is segmented into memory modules of uniform size, which are referred to as banks. Each bank occupies 4 bytes, and the continuous 128 bytes in the shared memory form 32 banks. Threads within a warp can access data from various banks simultaneously in one single memory access request. However, if multiple threads access data within the same bank concurrently, it will give rise to a bank conflict, leading to lower throughput. 

The above Figure~\ref{fig:bank-conflict} shows the scenario where bank conflict occurs (BM=8, BK=512), where E denotes an \textbf{e}lement of int32, B denotes \textbf{b}ank, and T denotes a \textbf{t}hread. When data is loaded from global memory to shared memory, each thread accesses 16 bytes in order to fully utilize the memory bandwidth. The 8x512-bit data is divided into four phases, each reading 128 bytes (i.e., two consecutive rows of data in the figure). During this process, no Bank conflicts occur in any of the phases.
However, when data is loaded from shared memory to registers, each thread needs to copy 4 bytes of data since the 8x8x128 BMMA stores matrix A (8x128 bits) and matrix B (8x128 bits) in 32-bit registers. When copying a matrix block (same color block), threads T0\~{}3, T8\~{}11, T16\~{}19, and T24\~{}27 all access B0\~{}3, while T4\~{}7, T12\~{}15, T20\~{}23, and T28\~{}31 access B16\~{}19, resulting in 4-way bank conflicts.

As depicted in the Figure~\ref{fig:bank-conflict-solution}, we employ the swizzle operation to address this issue. When the global memory is loaded into the shared memory, the addresses are swizzled, causing the 8x128bit data to be scattered across 32 banks. When T0\~{}31 accesses 8x128bit data, the data comes from B0\~{}31 respectively, and no bank conflict will arise, thereby further improving the efficiency of memory access.

\begin{table}[!ht] 
\centering
\resizebox{\columnwidth}{!}{%
\begin{tabular}{ccccccccc}
\hline
\multicolumn{9}{c}{LLaMA-7B} \\ \hline
 & \multicolumn{2}{c}{128} & \multicolumn{2}{c}{256} & \multicolumn{2}{c}{512} & \multicolumn{2}{c}{1024} \\ \cline{2-9} 
\multirow{-2}{*}{sequence length} & Latency(ms) & Memory(GB) & Latency(ms) & Memory(GB) & Latency(ms) & Memory(GB) & Latency(ms) & Memory(GB) \\ \hline
FP 16 & 1490.5 & 13.47 & 3005.95 & 13.534 & 6090.97 & 13.662 & 12561.82 & 13.918 \\
W8A16(CUTLASS) & 868.35 & 7.394 & 1755.62 & 7.458 & 3594.95 & 7.586 & 7559.22 & 7.842 \\
W8A8(SmoothQuant) & 832.25 & 7.394 & 1684.85 & 7.458 & 3445.86 & 7.586 & 7257.02 & 7.842 \\
W4A16(CUTLASS) & 642.24 & 4.258 & 1312.91 & 4.322 & 2707.26 & 4.45 & 5786.8 & 4.706 \\
W2A8(ABQ-LLM) & \textbf{505.60} & \textbf{2.784} & \textbf{1041.39} & \textbf{2.848} & \textbf{2167.30} & \textbf{2.976} & \textbf{4657.18} & \textbf{3.232} \\ \hline
\multicolumn{9}{c}{LLaMA-13B} \\ \hline
 & \multicolumn{2}{c}{128} & \multicolumn{2}{c}{256} & \multicolumn{2}{c}{512} & \multicolumn{2}{c}{1024} \\ \cline{2-9} 
\multirow{-2}{*}{sequence length} & Latency(ms) & Memory(GB) & Latency(ms) & Memory(GB) & Latency(ms) & Memory(GB) & Latency(ms) & Memory(GB) \\ \hline
FP 16 & 2726.66 & 25.6 & 5481.96 & 25.696 & 11071.81 & 25.92 & 22559.77 & 26.304 \\
W8A16(CUTLASS) & 1439.66 & 13.524 & 2900.46 & 13.62 & 5922.28 & 13.844 & 12257.27 & 14.228 \\
W8A8(SmoothQuant) & 1431.2 & 13.526 & 2874.51 & 13.622 & 5867.42 & 13.846 & 12193.68 & 14.23 \\
W4A16(CUTLASS) & 999.1 & 7.444 & 2020.99 & 7.54 & 4155.94 & 7.764 & 8750.98 & 8.148 \\
W2A8(ABQ-LLM) & \textbf{766.28} & \textbf{4.564} & \textbf{1550.21} & \textbf{4.66} & \textbf{3234.67} & \textbf{4.884} & \textbf{6885.69} & \textbf{5.268} \\ \hline
\multicolumn{9}{c}{LLaMA-30B} \\ \hline
 & \multicolumn{2}{c}{128} & \multicolumn{2}{c}{256} & \multicolumn{2}{c}{512} & \multicolumn{2}{c}{1024} \\ \cline{2-9} 
\multirow{-2}{*}{sequence length} & Latency(ms) & Memory(GB) & Latency(ms) & Memory(GB) & Latency(ms) & Memory(GB) & Latency(ms) & Memory(GB) \\ \hline
FP 16 & 3759.08 & 65.534 & 7540.17 & 65.726 & 15241.36 & 66.11 & 31073.23 &  66.878 \\
W8A16(CUTLASS) & 3032.64 & 32.418 & 6111.43 & 32.642 & 12371.66 & 33.026 & 25477.58 & 33.794 \\
W8A8(SmoothQuant) & 3057.96 & 32.418 & 6155.38 & 32.642 & 12465.96 & 33.026 & 25678.66 & 33.794 \\
W4A16(CUTLASS) & 1938.2 & 17.178 & 3924.2 & 17.402 & 8011.57 & 17.786 & 16680.64 & 18.554 \\
W2A8(ABQ-LLM) & \textbf{1730.51} & \textbf{9.616} & \textbf{3473.66} & \textbf{9.84} & \textbf{7133.13} & \textbf{10.224} & \textbf{14845.67} & \textbf{10.992} \\ \hline
\end{tabular}
}
\caption{Inference latency and memory usage of the FastTransformer implementation on NVIDIA A800-40GB GPU with a fixed input sequence length of 15, output sequence lengths of 128, 256, 512 and 1024. }
\label{tab:e2e}
\end{table}

\begin{table}[!ht]
\centering
\resizebox{\columnwidth}{!}{%
\begin{tabular}{cccccccccccccc}
\hline
\multicolumn{14}{c}{(1,4096)x(4096,4096)} \\ \hline
 &  & w2a2 & w2a4 & w2a6 & w2a8 & w3a3 & w3a8 & w4a4 & w4a8 & w5a5 & w6a6 & w7a7 & w8a8 \\
 & Ours(TOPS) & 5.126408 & 5.003512 & 4.993599 & 5.016534 & 1.844438 & 1.820546 & 1.415388 & 1.409558 & 1.167617 & 0.990153 & 0.853578 & 0.760755 \\
 & CUTLASS(TOPS) & 1.395987 & 1.395987 & 0.671532 & 0.671532 & 1.395987 & 0.671532 & 1.395987 & 0.671532 & 0.671532 & 0.671532 & 0.671532 & 0.671532 \\
\multirow{-3}{*}{RTX3070} & CUBLAS(TOPS) & - & - & 0.662154 & 0.662154 & - & 0.662154 & - & 0.662154 & 0.662154 & 0.662154 & 0.662154 & 0.662154 \\ \hline
\multicolumn{14}{c}{(1,1024)x(1024,8192)} \\ \hline
 & Ours(TOPS) & {\color[HTML]{1F2329} 3.959401} & {\color[HTML]{1F2329} 3.815584} & {\color[HTML]{1F2329} 3.710145} & {\color[HTML]{1F2329} 3.636041} & {\color[HTML]{1F2329} 3.27549} & {\color[HTML]{1F2329} 3.088989} & {\color[HTML]{1F2329} 2.729302} & {\color[HTML]{1F2329} 2.493001} & {\color[HTML]{1F2329} 1.107551} & {\color[HTML]{1F2329} 0.907098} & {\color[HTML]{1F2329} 0.774627} & {\color[HTML]{1F2329} 0.699006} \\
 & CUTLASS(TOPS) & 2.43737 & 2.43737 & 0.570752 & 0.570752 & 2.43737 & 0.570752 & 2.43737 & 0.570752 & 0.570752 & 0.570752 & 0.570752 & 0.570752 \\
\multirow{-3}{*}{RTX3070} & CUBLAS(TOPS) & - & - & 0.547283 & 0.547283 & - & 0.547283 & - & 0.547283 & 0.547283 & 0.547283 & 0.547283 & 0.547283 \\ \hline
\multicolumn{14}{c}{(1,11008)x(11008, 4096)} \\ \hline
 &  & w2a2 & w2a4 & w2a6 & w2a8 & w3a3 & w3a8 & w4a4 & w4a8 & w5a5 & w6a6 & w7a7 & w8a8 \\
 & Ours(TOPS) & {\color[HTML]{1F2329} 2.831276} & {\color[HTML]{1F2329} 2.845823} & {\color[HTML]{1F2329} 2.851906} & {\color[HTML]{1F2329} 2.85108} & {\color[HTML]{1F2329} 1.893524} & {\color[HTML]{1F2329} 1.89145} & {\color[HTML]{1F2329} 1.429216} & {\color[HTML]{1F2329} 1.424223} & {\color[HTML]{1F2329} 1.063806} & {\color[HTML]{1F2329} 0.9578} & {\color[HTML]{1F2329} 0.781478} & {\color[HTML]{1F2329} 0.688792} \\
 & CUTLASS(TOPS) & 1.437404 & 1.437404 & 0.702909 & 0.702909 & 1.437404 & 0.702909 & 1.437404 & 0.702909 & 0.702909 & 0.702909 & 0.702909 & 0.702909 \\
\multirow{-3}{*}{RTX3070} & CUBLAS(TOPS) & - & - & 0.655704 & 0.655704 & - & 0.655704 & - & 0.655704 & 0.655704 & 0.655704 & 0.655704 & 0.655704 \\ \hline
\multicolumn{14}{c}{(1,5120)x(5120, 5120)} \\ \hline
 &  & w2a2 & w2a4 & w2a6 & w2a8 & w3a3 & w3a8 & w4a4 & w4a8 & w5a5 & w6a6 & w7a7 & w8a8 \\
 & Ours(TOPS) & 2.77896 & 2.744274 & 2.769214 & 2.72645 & 1.946023 & 1.925246 & 1.485292 & 1.47769 & 1.207433 & 1.019555 & 0.88083 & 0.777159 \\
 & CUTLASS(TOPS) & 1.456822 & 1.456822 & 0.752888 & 0.752888 & 1.456822 & 0.752888 & 1.456822 & 0.752888 & 0.752888 & 0.752888 & 0.752888 & 0.752888 \\
\multirow{-3}{*}{RTX3070} & CUBLAS(TOPS) & - & - & 0.655704 & 0.655704 & - & 0.655704 & - & 0.655704 & 0.655704 & 0.655704 & 0.655704 & 0.655704 \\ \hline
\multicolumn{14}{c}{(1,4096)x(4096, 11008)} \\ \hline
 &  & w2a2 & w2a4 & w2a6 & w2a8 & w3a3 & w3a8 & w4a4 & w4a8 & w5a5 & w6a6 & w7a7 & w8a8 \\
 & Ours(TOPS) & 2.93919 & 2.945875 & 2.932437 & 2.928438 & 2.048219 & 2.042158 & 1.558931 & 1.556203 & 1.263128 & 1.06387 & 0.915512 & 0.804458 \\
 & CUTLASS(TOPS) & 1.44785 & 1.44785 & 0.757006 & 0.757006 & 1.44785 & 0.757006 & 1.44785 & 0.757006 & 0.757006 & 0.757006 & 0.757006 & 0.757006 \\
\multirow{-3}{*}{RTX3070} & CUBLAS(TOPS) & - & - & 0.726139 & 0.726139 & - &  & - & 0.726139 & 0.726139 & 0.726139 & 0.726139 & 0.726139 \\ \hline
\multicolumn{14}{c}{(4,4096)x(4096,4096)} \\ \hline
 &  & w2a2 & w2a4 & w2a6 & w2a8 & w3a3 & w3a8 & w4a4 & w4a8 & w5a5 & w6a6 & w7a7 & w8a8 \\
 & Ours(TOPS) & 19.85037 & 18.573332 & 17.173864 & 15.596383 & 7.165928 & 7.040589 & 5.588709 & 5.465432 & 4.427809 & 3.841075 & 3.17604 & 2.952538 \\
 & CUTLASS(TOPS) & 5.029238 & 5.029238 & 2.600067 & 2.600067 & 5.029238 & 2.600067 & 5.029238 & 2.600067 & 2.600067 & 2.600067 & 2.600067 & 2.600067 \\
\multirow{-3}{*}{RTX3070} & CUBLAS(TOPS) & - & - & 2.623434 & 2.623434 & - & 2.623434 & - & 2.623434 & 2.623434 & 2.623434 & 2.623434 & 2.623434 \\ \hline
\multicolumn{14}{c}{(4,1024)x(1024,8192)} \\ \hline
 &  & w2a2 & w2a4 & w2a6 & w2a8 & w3a3 & w3a8 & w4a4 & w4a8 & w5a5 & w6a6 & w7a7 & w8a8 \\
 & Ours(TOPS) & {\color[HTML]{1F2329} 14.911489} & {\color[HTML]{1F2329} 13.707592} & {\color[HTML]{1F2329} 12.752676} & {\color[HTML]{1F2329} 12.651737} & {\color[HTML]{1F2329} 11.246954} & {\color[HTML]{1F2329} 10.272099} & {\color[HTML]{1F2329} 8.611968} & {\color[HTML]{1F2329} 7.994145} & {\color[HTML]{1F2329} 4.100613} & {\color[HTML]{1F2329} 3.487998} & {\color[HTML]{1F2329} 2.770024} & {\color[HTML]{1F2329} 2.715393} \\
 & CUTLASS(TOPS) & 7.790775 & 7.790775 & 2.292671 & 2.292671 & 2.292671 & 2.292671 & 7.790775 & 2.292671 & 2.292671 & 2.292671 & 2.292671 & 2.292671 \\
\multirow{-3}{*}{RTX3070} & CUBLAS(TOPS) & - & - & 2.652782 & 2.652782 & - & 2.652782 & - & 2.652782 & 2.652782 & 2.652782 & 2.652782 & 2.652782 \\ \hline
\multicolumn{14}{c}{(4,11008)x(11008, 4096)} \\ \hline
 &  & w2a2 & w2a4 & w2a6 & w2a8 & w3a3 & w3a8 & w4a4 & w4a8 & w5a5 & w6a6 & w7a7 & w8a8 \\
 & Ours(TOPS) & {\color[HTML]{1F2329} 11.352832} & {\color[HTML]{1F2329} 11.176699} & {\color[HTML]{1F2329} 10.902382} & {\color[HTML]{1F2329} 10.897325} & {\color[HTML]{1F2329} 7.418883} & {\color[HTML]{1F2329} 7.308671} & {\color[HTML]{1F2329} 5.631782} & {\color[HTML]{1F2329} 5.359304} & {\color[HTML]{1F2329} 4.201856} & {\color[HTML]{1F2329} 3.729234} & {\color[HTML]{1F2329} 2.658403} & {\color[HTML]{1F2329} 2.734334} \\
 & CUTLASS(TOPS) & 6.089864 & 6.089864 & 2.852206 & 2.852206 & 6.089864 & 2.852206 & 6.089864 & 2.852206 & 2.852206 & 2.852206 & 2.852206 & 2.852206 \\
\multirow{-3}{*}{RTX3070} & CUBLAS(TOPS) & - & - & 2.804943 & 2.804943 & - &  & - & 2.804943 & 2.804943 & 2.804943 & 2.804943 & 2.804943 \\ \hline
\multicolumn{14}{c}{(4,5120)x(5120, 5120)} \\ \hline
 &  & w2a2 & w2a4 & w2a6 & w2a8 & w3a3 & w3a8 & w4a4 & w4a8 & w5a5 & w6a6 & w7a7 & w8a8 \\
 & Ours(TOPS) & 11.023197 & 10.805677 & 10.686147 & 10.633989 & 7.62955 & 7.509258 & 5.885227 & 5.81653 & 4.70967 & 4.021154 & 3.319287 & 3.049934 \\
 & CUTLASS(TOPS) & 5.41928 & 5.41928 & 2.892696 & 2.892696 & 5.41928 & 2.892696 & 5.41928 & 2.892696 & 2.892696 & 2.892696 & 2.892696 & 2.892696 \\
\multirow{-3}{*}{RTX3070} & CUBLAS(TOPS) & - & - & 3.024624 & 3.024624 & - & 3.024624 & - & 3.024624 & 3.024624 & 3.024624 & 3.024624 & 3.024624 \\ \hline
\multicolumn{14}{c}{(4,4096)x(4096, 11008)} \\ \hline
 &  & w2a2 & w2a4 & w2a6 & w2a8 & w3a3 & w3a8 & w4a4 & w4a8 & w5a5 & w6a6 & w7a7 & w8a8 \\
 & Ours(TOPS) & 11.65253 & 11.47451 & 11.370432 & 11.249153 & 8.044027 & 7.938343 & 6.158429 & 6.063344 & 4.936044 & 4.181711 & 3.516512 & 3.152066 \\
 & CUTLASS(TOPS) & 5.755347 & 5.755347 & 2.913199 & 2.913199 & 5.755347 & 2.913199 & 5.755347 & 2.913199 & 2.913199 & 2.913199 & 2.913199 & 2.913199 \\
\multirow{-3}{*}{RTX3070} & CUBLAS(TOPS) & - & - & 3.150072 & 3.150072 & - & 3.150072 & - & 3.150072 & 3.150072 & 3.150072 & 3.150072 & 3.150072 \\ \hline
\multicolumn{14}{c}{(8,4096)x(4096,4096)} \\ \hline
 &  & w2a2 & w2a4 & w2a6 & w2a8 & w3a3 & w3a8 & w4a4 & w4a8 & w5a5 & w6a6 & w7a7 & w8a8 \\
 & Ours(TOPS) & 34.711864 & 29.218008 & 25.562555 & 23.108412 & 14.124731 & 13.319459 & 10.805161 & 10.304756 & 8.570437 & 7.295355 & 6.372617 & 5.418213 \\
 & CUTLASS(TOPS) & 10.443984 & 10.443984 & 5.328563 & 5.328563 & 10.443984 & 5.328563 & 10.443984 & 5.328563 & 5.328563 & 5.328563 & 5.328563 & 5.328563 \\
\multirow{-3}{*}{RTX3070} & \multicolumn{1}{l}{CUBLAS(TOPS)} & - & - & 5.136078 & 5.136078 & - & 5.136078 & - & 5.136078 & 5.136078 & 5.136078 & 5.136078 & 5.136078 \\ \hline
\multicolumn{14}{c}{(8,1024)x(1024,8192)} \\ \hline
 &  & w2a2 & w2a4 & w2a6 & w2a8 & w3a3 & w3a8 & w4a4 & w4a8 & w5a5 & w6a6 & w7a7 & w8a8 \\
 & Ours(TOPS) & 25.143295 & 24.340204 & 22.991055 & 17.975151 & 21.315987 & 15.096982 & 14.203727 & 10.091 & 8.056054 & 6.547737 & 5.476622 & 4.921968 \\
 & CUTLASS(TOPS) & 11.837081 & 11.837081 & 4.52688 & 4.52688 & 11.837081 & 4.52688 & 11.837081 & 4.52688 & 4.52688 & 4.52688 & 4.52688 & 4.52688 \\
\multirow{-3}{*}{RTX3070} & CUBLAS(TOPS) & - & - & 5.051333 & 5.051333 & - & 5.051333 & - & 5.051333 & 5.051333 & 5.051333 & 5.051333 & 5.051333 \\ \hline
\multicolumn{14}{c}{(8,11008)x(11008, 4096)} \\ \hline
 &  & w2a2 & w2a4 & w2a6 & w2a8 & w3a3 & w3a8 & w4a4 & w4a8 & w5a5 & w6a6 & w7a7 & w8a8 \\
 & Ours(TOPS) & 21.93853 & 21.763004 & 21.38822 & 21.141041 & 14.794146 & 14.292913 & 10.0646 & {\color[HTML]{1F2329} 9.831863} & 8.391643 & 6.331276 & 5.216482 & 5.386425 \\
 & CUTLASS(TOPS) & 11.559231 & 11.559231 & 5.788068 & 5.788068 & 11.559231 & 5.788068 & 11.559231 & 5.788068 & 5.788068 & 5.788068 & 5.788068 & 5.788068 \\
\multirow{-3}{*}{RTX3070} & CUBLAS(TOPS) & - & - & 5.433702 & 5.433702 & - & 5.433702 & - & 5.433702 & 5.433702 & 5.433702 & 5.433702 & 5.433702 \\ \hline
\multicolumn{14}{c}{(8,5120)x(5120, 5120)} \\ \hline
 &  & w2a2 & w2a4 & w2a6 & w2a8 & w3a3 & w3a8 & w4a4 & w4a8 & w5a5 & w6a6 & w7a7 & w8a8 \\
 & Ours(TOPS) & 21.438643 & 20.998667 & 20.403486 & 20.144592 & 15.058184 & 14.257557 & 11.512318 & 11.014898 & 9.303804 & 7.909779 & 6.563432 & 5.816199 \\
 & CUTLASS(TOPS) & 11.438784 & 11.438784 & 5.650824 & 5.650824 & 11.438784 & 5.650824 & 11.438784 & 5.650824 & 5.650824 & 5.650824 & 5.650824 & 5.650824 \\
\multirow{-3}{*}{RTX3070} & CUBLAS(TOPS) & - & - & 5.843657 & 5.843657 & - & 5.843657 & - & 5.843657 & 5.843657 & 5.843657 & 5.843657 & 5.843657 \\ \hline
\multicolumn{14}{c}{(8,4096)x(4096, 11008)} \\ \hline
 &  & w2a2 & w2a4 & w2a6 & w2a8 & w3a3 & w3a8 & w4a4 & w4a8 & w5a5 & w6a6 & w7a7 & w8a8 \\
 & Ours(TOPS) & 22.735727 & 22.486816 & 21.946047 & 21.290138 & 15.927655 & 14.94954 & 12.047882 & 11.759899 & 9.831679 & 8.178876 & 6.951003 & 5.942341 \\
 & CUTLASS(TOPS) & 12.133122 & 12.133122 & 5.794113 & 5.794113 & 12.133122 & 5.794113 & 12.133122 & 5.794113 & 5.794113 & 5.794113 & 5.794113 & 5.794113 \\
\multirow{-3}{*}{RTX3070} & CUBLAS(TOPS) & - & - & 6.094711 & 6.094711 & - & 6.094711 & - & 6.094711 & 6.094711 & 6.094711 & 6.094711 & 6.094711 \\ \hline
\end{tabular}
}
\caption{The GEMM speed comparison of our ABQKernel, CUTLASS, and cuBLAS in RTX 3070. }
\label{tab:3070}
\end{table}

\begin{table}[!th]
\centering
\resizebox{\columnwidth}{!}{%
\begin{tabular}{cccccccccccccc}
\toprule
\multicolumn{14}{c}{(1,4096)x(4096,4096)} \\ \midrule
 &  & w2a2 & w2a4 & w2a6 & w2a8 & w3a3 & w3a8 & w4a4 & w4a8 & w5a5 & w6a6 & w7a7 & w8a8 \\
 & Ours(TOPS) & 5.666263 & 5.596584 & 5.177539 & 5.192204 & 4.932711 & 5.103254 & 5.230327 & 3.984436 & 5.089779 & 4.697921 & 2.73454 & 3.997423 \\
 & CUTLASS(TOPS) & 4.378407 & 4.378407 & 2.483177 & 2.483177 & 4.378407 & 2.483177 & 4.378407 & 2.483177 & 2.483177 & 2.483177 & 2.483177 & 2.483177 \\
\multirow{-3}{*}{RTX4080} & CUBLAS(TOPS) & - & - & 2.132917 & 2.132917 & - & 2.132917 & - & 2.132917 & 2.132917 & 2.132917 & 2.132917 & 2.132917 \\ \midrule
\multicolumn{14}{c}{(1,1024)x(1024,8192)} \\ \midrule
 &  & w2a2 & w2a4 & w2a6 & w2a8 & w3a3 & w3a8 & w4a4 & w4a8 & w5a5 & w6a6 & w7a7 & w8a8 \\
 & Ours(TOPS) & 3.544786 & 3.494135 & 3.491157 & 3.335505 &  3.090738 & 3.463115 & 3.280737 &  2.427181 &  2.593226 &  2.545285 & 2.018674 & 2.512498 \\
 & CUTLASS(TOPS) & 1.994158 & 1.994158 & 1.595792 & 1.595792 & 1.994158 & 1.595792 & 1.994158 & 1.595792 & 1.595792 & 1.595792 & 1.595792 & 1.595792 \\
\multirow{-3}{*}{RTX4080} & CUBLAS(TOPS) & - & - & 1.842555 & 1.842555 & - & 1.842555 & - & 1.842555 & 1.842555 & 1.842555 & 1.842555 & 1.842555 \\ \midrule
\multicolumn{14}{c}{(1,11008)x(11008, 4096)} \\ \midrule
 &  & w2a2 & w2a4 & w2a6 & w2a8 & w3a3 & w3a8 & w4a4 & w4a8 & w5a5 & w6a6 & w7a7 & w8a8 \\
 & Ours(TOPS) & 11.459206 & 11.172798 &  8.219525 & 8.073364 & 6.89011 & 8.299166 & 7.945147 & 5.589236 & 6.926538 & 6.253657 &  3.5678 &  5.010184 \\
 & CUTLASS(TOPS) & 5.738564 & 5.738564 & 3.577838 & 3.577838 & 5.738564 & 3.577838 & 5.738564 & 3.577838 & 3.577838 & 3.577838 & 3.577838 & 3.577838 \\
\multirow{-3}{*}{RTX4080} & CUBLAS(TOPS) & - & - & 2.782205 & 2.782205 & - & 2.782205 & - & 2.782205 & 2.782205 & 2.782205 & 2.782205 & 2.782205 \\ \midrule
\multicolumn{14}{c}{(1,5120)x(5120, 5120)} \\ \midrule
 &  & w2a2 & w2a4 & w2a6 & w2a8 & w3a3 & w3a8 & w4a4 & w4a8 & w5a5 & w6a6 & w7a7 & w8a8 \\
 & Ours(TOPS) & 8.179976 & 8.036415 & 7.981294 & 7.326846 & 7.348096 & 7.640651 & 7.2295 & 4.169042 & 6.045578 & 4.925341 & 3.0402 & 4.270938 \\
 & CUTLASS(TOPS) & 4.469664 & 4.469664 & 2.111863 & 2.111863 & 4.469664 & 2.111863 & 4.469664 & 2.111863 & 2.111863 & 2.111863 & 2.111863 & 2.111863 \\
\multirow{-3}{*}{RTX4080} & CUBLAS(TOPS) & - & - & 2.338113 & 2.338113 & - &  & - & 2.338113 & 2.338113 & 2.338113 & 2.338113 & 2.338113 \\ \midrule
\multicolumn{14}{c}{(1,4096)x(4096, 11008)} \\ \midrule
 &  & w2a2 & w2a4 & w2a6 & w2a8 & w3a3 & w3a8 & w4a4 & w4a8 & w5a5 & w6a6 & w7a7 & w8a8 \\
 & Ours(TOPS) & 13.542058 & 13.110614 & 12.246419 & 9.048911 & 7.191247 & 10.480065 & 7.804325 & 4.767432 & 6.815571 & 5.946993 & 3.366875 & 5.155972 \\
 & CUTLASS(TOPS) & 7.373691 & 7.373691 & 3.04093 & 3.04093 & 7.373691 & 3.04093 & 7.373691 & 3.04093 & 3.04093 & 3.04093 & 3.04093 & 3.04093 \\
\multirow{-3}{*}{RTX4080} & CUBLAS(TOPS) & - & - & 2.862102 & 2.862102 & - & 2.862102 & - & 2.862102 & 2.862102 & 2.862102 & 2.862102 & 2.862102 \\ \midrule
\multicolumn{14}{c}{(4,4096)x(4096,4096)} \\ \midrule
 &  & w2a2 & w2a4 & w2a6 & w2a8 & w3a3 & w3a8 & w4a4 & w4a8 & w5a5 & w6a6 & w7a7 & w8a8 \\
 & Ours(TOPS) & 24.408194 & 20.693401 & 20.195993 & 20.795177 & 18.665907 & 20.496012 & 19.760592 & 12.737339 & 18.525913 & 16.639837 & 8.414879 & 11.580844 \\
 & CUTLASS(TOPS) & 16.965191 & 16.965191 & 11.346475 & 11.346475 & 16.965191 & 11.346475 & 16.965191 & 11.346475 & 11.346475 & 11.346475 & 11.346475 & 11.346475 \\
\multirow{-3}{*}{RTX4080} & CUBLAS(TOPS) & - & - & 9.346264 & 9.346264 & - & 9.346264 & - & 9.346264 & 9.346264 & 9.346264 & 9.346264 & 9.346264 \\ \midrule
\multicolumn{14}{c}{(4,1024)x(1024,8192)} \\ \midrule
 &  & w2a2 & w2a4 & w2a6 & w2a8 & w3a3 & w3a8 & w4a4 & w4a8 & w5a5 & w6a6 & w7a7 & w8a8 \\
 & Ours(TOPS) & 14.124137 & 14.265563 &  14.515171 & 13.744966 & 13.02943 & 12.581451 & 11.312963 & 9.696109 & 10.008553 &10.04691 &  5.635082 & 8.187905 \\
 & CUTLASS(TOPS) & 10.125006 & 10.125006 & 7.514869 & 7.514869 & 10.125006 & 7.514869 & 10.125006 & 7.514869 & 7.514869 & 7.514869 & 7.514869 & 7.514869 \\
\multirow{-3}{*}{RTX4080} & CUBLAS(TOPS) & - & - & 5.807355 & 5.807355 & - & 5.807355 & - & 5.807355 & 5.807355 & 5.807355 & 5.807355 & 5.807355 \\ \midrule
\multicolumn{14}{c}{(4,11008)x(11008, 4096)} \\ \midrule
 &  & w2a2 & w2a4 & w2a6 & w2a8 & w3a3 & w3a8 & w4a4 & w4a8 & w5a5 & w6a6 & w7a7 & w8a8 \\
 & Ours(TOPS) & 44.499241 & 34.592556 &  31.936174 &  30.138264 & 26.151152 &  29.124102 & 30.114992 & 20.55409 & 25.744062 & 21.976168 & 11.041816 &16.608015 \\
 & CUTLASS(TOPS) & 22.360314 & 22.360314 & 14.27062 & 14.27062 & 22.360314 & 14.27062 & 22.360314 & 14.27062 & 14.27062 & 14.27062 & 14.27062 & 14.27062 \\
\multirow{-3}{*}{RTX4080} & CUBLAS(TOPS) & - & - & 10.484434 & 10.484434 & - & 10.484434 & - & 10.484434 & 10.484434 & 10.484434 & 10.484434 & 10.484434 \\ \midrule
 &  & \multicolumn{12}{c}{(4,5120)x(5120, 5120)} \\ \midrule
 &  & w2a2 & w2a4 & w2a6 & w2a8 & w3a3 & w3a8 & w4a4 & w4a8 & w5a5 & w6a6 & w7a7 & w8a8 \\
 & Ours(TOPS) & 32.262131 & 32.364094 & 31.850698 & 26.226151 & 20.416708 & 24.156643 & 26.690235 & 14.489883 & 18.306963 & 18.400719 & 8.766748 & 13.434794 \\
 & CUTLASS(TOPS) & 15.777285 & 15.777285 & 8.469383 & 8.469383 & 15.777285 & 8.469383 & 15.777285 & 8.469383 & 8.469383 & 8.469383 & 8.469383 & 8.469383 \\
\multirow{-3}{*}{RTX4080} & CUBLAS(TOPS) & - & - & 6.160892 & 6.160892 & - & 6.160892 & - & 6.160892 & 6.160892 & 6.160892 & 6.160892 & 6.160892 \\ \midrule
\multicolumn{14}{c}{(4,4096)x(4096, 11008)} \\ \midrule
 &  & w2a2 & w2a4 & w2a6 & w2a8 & w3a3 & w3a8 & w4a4 & w4a8 & w5a5 & w6a6 & w7a7 & w8a8 \\
 & Ours(TOPS) & 52.217018 & 50.157482 & 45.968418 & 31.412338 & 28.61311 & 30.551258 & 30.689667 & 17.375622 & 22.56772 & 21.429615 & 10.303197 & 16.78546 \\
 & CUTLASS(TOPS) & 29.325113 & 29.325113 & 12.020953 & 12.020953 & 29.325113 & 12.020953 & 29.325113 & 12.020953 & 12.020953 & 12.020953 & 12.020953 & 12.020953 \\
\multirow{-3}{*}{RTX4080} & CUBLAS(TOPS) & - & - & 9.30319 & 9.30319 & - & 9.30319 & - & 9.30319 & 9.30319 & 9.30319 & 9.30319 & 9.30319 \\ \midrule
\multicolumn{14}{c}{(8,4096)x(4096,4096)} \\ \midrule
 &  & w2a2 & w2a4 & w2a6 & w2a8 & w3a3 & w3a8 & w4a4 & w4a8 & w5a5 & w6a6 & w7a7 & w8a8 \\
 & Ours(TOPS) & 41.439137 & 41.30854 & 41.419498 & 39.114296 & 37.503555 & 37.120361 & 40.611 & 24.405922 & 32.121555 & 24.77029 & 16.262033 & 17.331835 \\
 & CUTLASS(TOPS) & 33.269646 & 33.269646 & 22.390638 & 22.390638 & 33.269646 & 22.390638 & 33.269646 & 22.390638 & 22.390638 & 22.390638 & 22.390638 & 22.390638 \\
\multirow{-3}{*}{RTX4080} & \multicolumn{1}{l}{} & - & - & 18.537868 & 18.537868 & - & 18.537868 & - & 18.537868 & 18.537868 & 18.537868 & 18.537868 & 18.537868 \\ \midrule
 &  & \multicolumn{12}{c}{(8,1024)x(1024,8192)} \\ \midrule
 &  & w2a2 & w2a4 & w2a6 & w2a8 & w3a3 & w3a8 & w4a4 & w4a8 & w5a5 & w6a6 & w7a7 & w8a8 \\
 & Ours(TOPS) & 28.155737 & 27.346548 & 26.414156 & 20.811686 & 20.343021 & 20.249033 & 20.602325 & 18.68187 & 19.067791 & 18.388329 & 11.315893 & 11.722744 \\
 & CUTLASS(TOPS) & 16.941822 & 16.941822 & 11.685739 & 11.685739 & 16.941822 & 11.685739 & 16.941822 & 11.685739 & 11.685739 & 11.685739 & 11.685739 & 11.685739 \\
\multirow{-3}{*}{RTX4080} & CUBLAS(TOPS) & - & - & 11.626171 & 11.626171 & - & 11.626171 & - & 11.626171 & 11.626171 & 11.626171 & 11.626171 & 11.626171 \\ \midrule
\multicolumn{14}{c}{(8,11008)x(11008, 4096)} \\ \midrule
 &  & w2a2 & w2a4 & w2a6 & w2a8 & w3a3 & w3a8 & w4a4 & w4a8 & w5a5 & w6a6 & w7a7 & w8a8 \\
 & Ours(TOPS) & 70.776772 & 63.515327 & 62.89724 & 53.603973 & 50.18964 & 44.507675 & 52.819912 & 35.874935 & 41.623066 & 34.309536 & 20.632343 & 20.644436 \\
 & CUTLASS(TOPS) & 42.806332 & 42.806332 & 28.412611 & 28.412611 & 42.806332 & 28.412611 & 42.806332 & 28.412611 & 28.412611 & 28.412611 & 28.412611 & 28.412611 \\
\multirow{-3}{*}{RTX4080} & CUBLAS(TOPS) & - & - & 20.376932 & 20.376932 & - & 20.376932 & - & 20.376932 & 20.376932 & 20.376932 & 20.376932 & 20.376932 \\ \midrule
\multicolumn{14}{c}{(8,5120)x(5120, 5120)} \\ \midrule
 &  & w2a2 & w2a4 & w2a6 & w2a8 & w3a3 & w3a8 & w4a4 & w4a8 & w5a5 & w6a6 & w7a7 & w8a8 \\
 & Ours(TOPS) & 62.192532 & 58.058117 & 56.991051 & 38.496239 & 36.794827 & 36.71896 & 41.916618 & 27.445724 & 35.866901 & 26.239592 & 16.931917 & 20.035217 \\
 & CUTLASS(TOPS) & 35.998506 & 35.998506 & 16.818377 & 16.818377 & 35.998506 & 16.818377 & 35.998506 & 16.818377 & 16.818377 & 16.818377 & 16.818377 & 16.818377 \\
\multirow{-3}{*}{RTX4080} & CUBLAS(TOPS) & - & - & 11.887421 & 11.887421 & - & 11.887421 & - & 11.887421 & 11.887421 & 11.887421 & 11.887421 & 11.887421 \\ \midrule
\multicolumn{14}{c}{(8,4096)x(4096, 11008)} \\ \midrule
 &  & w2a2 & w2a4 & w2a6 & w2a8 & w3a3 & w3a8 & w4a4 & w4a8 & w5a5 & w6a6 & w7a7 & w8a8 \\
 & Ours(TOPS) & 103.437386 & 66.689888 & 63.139633 & 49.792355 & 51.081207 & 50.856277 & 52.395653 & 33.181564 & 43.949936 & 34.850956 & 19.520435 & 21.211935 \\
 & CUTLASS(TOPS) & 55.537205 & 55.537205 & 24.016449 & 24.016449 & 55.537205 & 24.016449 & 55.537205 & 24.016449 & 24.016449 & 24.016449 & 24.016449 & 24.016449 \\
\multirow{-3}{*}{RTX4080} & CUBLAS(TOPS) & - & - & 18.105219 & 18.105219 & - &  & - & 18.105219 & 18.105219 & 18.105219 & 18.105219 & 18.105219 \\ \bottomrule
\end{tabular}%
}
\caption{The GEMM speed comparison of our ABQKernel, CUTLASS, and cuBLAS in RTX 4080. }
\label{tab:4080}
\end{table}

\end{document}